\documentclass[manuscript]{acmart}
\setcopyright{acmcopyright}
\copyrightyear{2018}
\acmYear{2018}
\acmDOI{XXXXXXX.XXXXXXX}

\acmConference[Conference acronym 'XX]{Make sure to enter the correct
  conference title from your rights confirmation emai}{June 03--05,
  2018}{Woodstock, NY}

\acmBooktitle{Woodstock '18: ACM Symposium on Neural Gaze Detection,
 June 03--05, 2018, Woodstock, NY} 
\acmPrice{15.00}
\acmISBN{978-1-4503-XXXX-X/18/06}

\usepackage[frozencache,cachedir=.]{minted}
\usepackage{fancyvrb}
\usepackage{graphicx}
\usepackage{pythonhighlight}
\usepackage{booktabs}  
\usepackage{caption}
\usepackage{subcaption}
\usepackage{csquotes}
\usepackage{framed}
\usepackage[normalem]{ulem} 
\usepackage{listings}
\usepackage{soul,xcolor}
\usepackage{mdframed}
\usepackage{lipsum} 
\newsavebox{\measurebox}
\usepackage[export]{adjustbox}
\usepackage{multirow}
\usepackage{colortbl}

\definecolor{colorins}{RGB}{18, 166, 131}
\definecolor{colordel}{RGB}{227, 30, 62}
\definecolor{slackbrown}{RGB}{63, 15, 64}
\definecolor{colorcomment}{RGB}{254, 241, 216}

\DeclareRobustCommand{\hlcomment}[1]{{\sethlcolor{colorcomment}\hl{#1}}}

\newcommand{\symbolimg}[2][0.4cm]{%
  \includegraphics[height=#1,valign=c]{#2}%
}

\newenvironment{myjsonblock}[1]{%
    \begin{center}
    \begin{minipage}{0.5\textwidth}
    \begin{mdframed}[
        linecolor=black,
        linewidth=1pt,
        innertopmargin=6pt,
        innerbottommargin=6pt,
        innerrightmargin=10pt,
        innerleftmargin=10pt,
        frametitle=#1,
        frametitlebackgroundcolor=slackbrown, 
        frametitlefont=\color{white}\bfseries, 
    ]
}{%
    \end{mdframed}
    \end{minipage}
    \end{center}
}

\begin{document}

\title{Beyond the Chat: Executable and Verifiable Text-Editing with LLMs}

\author{Philippe Laban}
\email{plaban@salesforce.com}
\affiliation{
  \institution{Salesforce AI Research}
  \country{United States}
}
\author{Jesse Vig}
\affiliation{
  \institution{Salesforce AI Research}
  \country{United States}
}
\author{Marti A. Hearst}
\affiliation{
  \institution{UC Berkeley}
  \country{United States}
}
\author{Caiming Xiong}
\affiliation{
  \institution{Salesforce AI Research}
  \country{United States}
}
\author{Chien-Sheng Wu}
\affiliation{
  \institution{Salesforce AI Research}
  \country{United States}
}

\begin{abstract}
Conversational interfaces powered by Large Language Models (LLMs) have recently become a popular way to obtain feedback during document editing. However, standard chat-based conversational interfaces do not support transparency and verifiability of the editing changes that they suggest. To give the author more agency when editing with an LLM, we present InkSync, an editing interface that suggests executable edits directly within the document being edited. Because LLMs are known to introduce factual errors, Inksync also supports a 3-stage approach to mitigate this risk: Warn authors when a suggested edit introduces new information, help authors Verify the new information's accuracy through external search, and allow an auditor to perform an a-posteriori verification by Auditing the document via a trace of all auto-generated content.
Two usability studies confirm the effectiveness of InkSync's components when compared to standard LLM-based chat interfaces, leading to more accurate, more efficient editing, and improved user experience.
\end{abstract}




\begin{teaserfigure}
    \centering
    \includegraphics[width=0.85\textwidth]{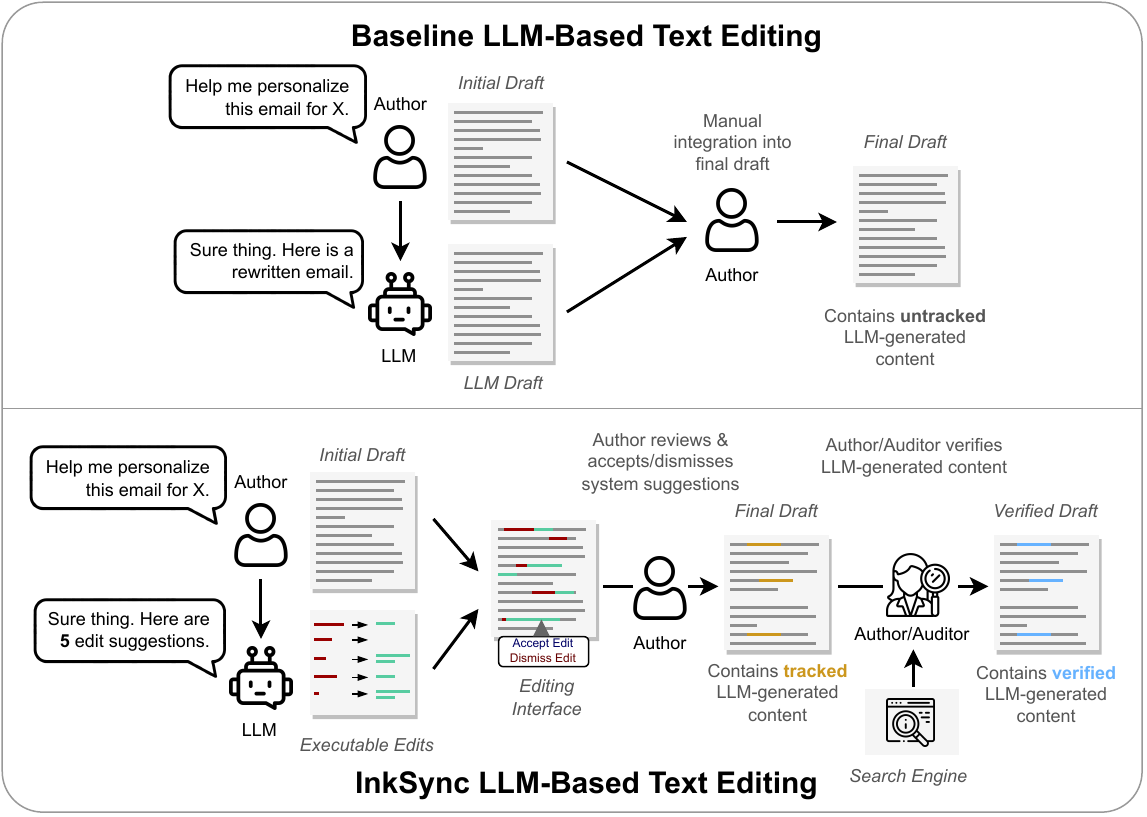}
    \caption{We design and evaluate the InkSync text editing interface, which relies on an LLM to suggest executable edits that can be verified and accepted into a draft. A paired audit interface enables the tracking of auto-generated content in a document's latest version and the verification of the factual accuracy of such content.} 
    \label{fig:intro_figure}
\end{teaserfigure}


\maketitle

\section{Introduction}

For many years, authors have used computer-assisted editing tools to improve their writing, as seen in the grammar suggestions of commercial software like Microsoft Word, Google Docs, and Grammarly. With the recent advances in Large Language Models (LLMs), the variety and power of automated suggestions for editing and rewriting has greatly increased. LLMs can go beyond typo and grammar suggestions, providing the ability to improve or change the tone (e.g., ``make this more informal''), personalize the text for a target reader (e.g., ``rewrite for a family with two children''), and suggest new content (e.g., ``suggest attractions for the family to visit''). 

Editing is a popular use of LLM-based chatbots. OpenAI's GPT3 API statistics \cite{ouyang2022training} reveal that when breaking down use cases by task type, ``rewrite'' tasks make up 6.6\% of all queries (even more than summarization queries at 4.2\%), implying the popularity of LLM-based feedback when editing text documents.

Yet standard LLM conversational interaction interfaces such as the popular ChatGPT lack several desirable properties. As illustrated in Figure~\ref{fig:intro_figure} (top), in a conversational interface, the LLM typically replies to a user's rewrite query with a fully revised draft.
This puts the burden on the author to mentally compare their draft and the LLM's proposed draft to assess changes. 
Authors must either manually integrate the suggestions, or be tempted to use the entire LLM-generated revision. 
We conducted a survey of professional workers (Section~\ref{sec:survey}) that found that 80\% of respondents use chat-based LLMs for editing documents, and while most (86\%) manually integrate the suggestions into their documents, 17\% copy the new version of the document verbatim.

To remedy these limitations, we introduce a new tool called InkSync for incorporating LLM-generated suggestions into the editing process. As shown in Figure~\ref{fig:intro_figure} (bottom), this tool introduces a novel editing workflow, allowing the author to clearly see which edits have been suggested in context, accept or reject those edits, view warnings about which suggested content contains new information, and scaffold the verification of that new information. 
InkSync incorporates four LLM-based components (Chat, Comment, Markers, and Brainstorm; Section~\ref{sec:system}) that can generate an explicit list of \textit{executable edits}, which are displayed to the author directly within the text editing interface. The author can then easily accept or dismiss the suggested edits based on their judgment and intended goals.

We conduct two empirical studies to assess the use of Inksync to produce better and more accurate text than a standard chat-based LLM alone.
 
In the first study (Section~\ref{sec:study1}), we explore the usability of the four edit-suggesting components and their relative effectiveness in helping participants achieve two objectives: (1) improving the tone and grammar of an email, and (2) customizing the email to a specific recipient.
Our findings reveal that participants using interfaces with these editing components are more successful at achieving both objectives than manual editing or a standard chat-based interface alone. We also find that the LLM-enabled components can complement each other: participants using an interface with all four editing components were the most effective at accomplishing the editing objectives.

An important and well-known limitation of using LLM-generated content in practical applications is the absence of a factual accuracy guarantee, as LLMs are known to hallucinate (make up) facts \cite{bang2023multitask,kryscinski2019neural}.
In an analysis conducted on 200 documents edited during the first study, we found that 28\% of the documents contained minor factual errors, and 12\% contained major factual errors, confirming that LLM's factual accuracy limitations extend to the domain of text editing.
To directly address the factual accuracy limitation of LLM-based interaction, we implement a three-stage approach in the InkSync interface: Warn, Verify, and Audit. 

\begin{itemize}
\setlength{\itemindent}{.8in}
\item[Stage 1: Warn:] The system shows the author a visual warning for suggested edits that would \textit{add new information} to the document,
\item[Stage 2: Verify:] For edits that introduce new information, the author is given the option to accept, dismiss, or \textit{verify} the edit, with a list of automatically-generated search engine queries to aid in verification.
\item[Stage 3: Audit:] 
An \textit{auditing interface} can be used to trace all system-generated content present in the latest document version, allowing an auditor to verify the accuracy of any LLM suggestion in the product of the editing process.
\end{itemize}

In the second study (Section~\ref{sec:study2}), we focus on evaluating the Warn, Verify, and Audit components of InkSync. Our findings indicate that a majority of participants find the warnings useful in deciding which edits to verify, that the verification procedure we propose helps reduce the number of inaccurate edits accepted into the document by 44\%, and that the auditing stage can further help identify up to 73\% of inaccurate edits, playing a complementary role to edit-time verification.

In summary, our contributions are the following:
\begin{itemize}
    \item We design and implement the InkSync system, a text editing interface backed by four LLM-based components -- Markers, Chat, Comment, and Brainstorm -- that suggest executable edits to the user, providing transparent changes and enabling traceability of LLM content during the editing process,
    \item We design and implement a three-stage Warn, Verify, and Audit approach to empower the user to know when the system is at risk of introducing factually inaccurate content, and to aid the user in verifying and auditing LLM content either at edit time or in a later audit,
    \item We conduct two usability studies that experimentally validate the effectiveness of the InkSync system for the task of editing and customizing outreach emails, and that the Warn, Verify and Audit approach is an effective human-in-the-loop method to allow for advanced LLM-based text editing while mitigating the risk of introducing factually inaccurate content.
\end{itemize}

We plan to open-source the InkSync system -- which is LLM agnostic -- and the corpus of edited documents from our usability studies, to enable research on the future of LLM-based text editing interfaces.

\section{Related Work} \label{sec:related_work}

\subsection{Intelligent Text-Editing Interfaces} \label{sec:rel_work_interfaces}

Before LLMs, prior work looked at crowd workers as a source of intelligence to assist in writing tasks, with systems such as Soylent \cite{bernstein2010soylent}, MicroWriter \cite{teevan2016supporting}, and WearWrite \cite{nebeling2016wearwrite} designing various crowd-source editing interfaces, with key considerations for task decomposition, cost, and latency. Since then, recent work has drawn a parallel between LLMs and crowdwork pipelines \cite{wu2023llms}.

Different suggestion mechanisms were proposed in prior work on co-writing with a model. Most commonly, the model is used for text-completion (i.e., auto-complete) \cite{coenen2021wordcraft,lee2022coauthor,buschek2021impact,calderwood2020novelists}, which is compatible with the model's training objective \cite{wu2018smart}. Yet auto-complete has drawbacks: it leads to reduced perceived authorship in the user \cite{lehmann2022suggestion}, and better language modeling performance is not indicative of successful human-LM interaction \cite{lee2022evaluating}. Some other work has proposed other mechanisms such as infilling \cite{coenen2021wordcraft, ippolito2022case} (i.e., leaving a blank in the document, which the model fills in), generic paraphrasing functions \cite{coenen2021wordcraft}, or systems that provide high-level feedback in the form of questions to the author, but no executable suggestions \cite{kim2023repurposing}. In InkSync, we implemented four edit-suggesting components that share a common executable edit language. The components each provide interactions beyond auto-complete or infilling.

\subsection{Evaluation Tasks for Text Editing} \label{sec:rel_work_tasks}

Prior work has explored a diverse set of tasks to evaluate intelligent text-editing interfaces. Most commonly, studies have been designed on creative tasks like short-story writing \cite{calderwood2020novelists,coenen2021wordcraft,lee2022coauthor,chung2022talebrush,singh2022hide,clark2018creative}, poetry \cite{chakrabarty2022help}, slogans \cite{clark2018creative}, or screenplays \cite{mirowski2023co}. In such tasks, a central consideration is the model's impact on creativity, which is challenging to evaluate. Recent work has proposed expository writing -- in which an author reviews a document corpus, synthesizes it, and adds their interpretation -- as a conducive task in the evaluation of AI-assisted writing support tools \cite{shen2023beyond}. In expository writing tasks, the system requirements might go beyond standard editing suggestions, with the system potentially providing inspiration \cite{gero2022sparks}, retrieving documents \cite{han2022passages}, and organizing the content hierarchically \cite{kang2022threddy}, which adds complexity to the evaluated system. In our work, we focus on the task of customizing a sales email, for several reasons. First, the knowledge workers we recruit as participants report that they frequently write and edit emails. Second, the task requires creativity from the participants, who can rely on the system for specific suggestions while allowing us to set a quantitative evaluation to measure participant success.

\subsection{Ethical Considerations of Co-writing} \label{sec:rel_work_ethics}

Some prior work has discussed the ethical considerations of co-writing with a system, with issues of user control, authorship, or leadership \cite{chen2023next}. \citet{rezwana2022identifying} find that a communicative agent leads to an improved collaborative experience (which we implement in the Chat and Comment InkSync components). \citet{biermann2022tool} discuss the need for professional authors to retain control over their writing process, and \citet{draxler2023ai} discuss an AI Ghostwriter effect: although human writers do not claim ownership of system-generated content, they are reluctant to publicly declare AI authorship. Finally, \citet{lehmann2022suggestion} found that auto-complete reduces the sense of ownership compared to suggestion-based designs. In InkSync, the executable nature of system suggestions gives editing control to the author and automatic traceability of system-generated text provides the auditing tools facilitating discussions of ownership and authorship.

\subsection{Text Editing in NLP} \label{sec:rel_work_nlp}

Previous NLP work on text editing focuses on various text editing formulations tasks, such as news or Wikipedia simplification \cite{xu2015problems,laban2023swipe}, biased text neutralization \cite{pryzant2020automatically}, or sentence decontextualization \cite{choi2021decontextualization}. Some work has framed text editing as an iterative process \cite{du2022read} and collected datasets of multi-turn interaction such as IteraTER \cite{du2022understanding} or Coauthor \cite{lee2022coauthor}. In turn, prior work has proposed specialized models including PEER \cite{schick2022peer} and CoEdIT \cite{raheja2023coedit} that can provide iterative suggestions to a user. In InkSync, we adapt an LLM to dynamically and iteratively suggest executable edits.

NLP evaluation largely uses automatic metrics to measure system success, including n-gram metrics such as BLEU \cite{papineni2002bleu} or SARI \cite{xu2016optimizing}, or more advanced learned metrics such as LENS \cite{maddela2022lens}. While much work focuses on English, recent work includes multi-lingual benchmarks \cite{ryan2023revisiting}. Some work investigates human evaluation of text-editing models, usually via intrinsic evaluation using scoring scales \cite{alva2021suitability}, or ranking \cite{maddela2022lens} to judge the success of the edited text in achieving a desired objective \cite{laban2021keep}. As is common in NLP, interface usability and user experience are out of the scope of the evaluation.

\subsection{Information Verification \& Auditing}

Some prior work has explored the different stages of document editing, either through a design space of \cite{gero2022design} for writing support tools, or through the survey of authors on the lifespan of their documents \cite{sarrafzadeh2021characterizing}. Both works find a need for specialized interfaces to support each stage of writing, and suggest a common review phase that follows an editing phase. In InkSync, we separate editing and auditing phases through a specialized interface, with a novel focus on tracing content provenance and verifying the factual accuracy of system-generated content.

Factual verification is an active NLP research area, with two task configurations: closed- and open-domain. In closed-domain verification, such as inconsistency detection in summarization \cite{maynez2020faithfulness,kryscinski2019neural,honovich2022true,laban2023llms}, a model must assess whether system-generated content (i.e., a summary) is consistent with provided source content (i.e., a document), and multiple modeling solutions have been proposed \cite{laban2022summac,fabbri2022qafacteval,goyal2021annotating}. Open-domain verification is more challenging, as it requires retrieving relevant documents and recording evidence sentences, with a community centering around benchmarks such as FEVER \cite{thorne2018fever} and proposing methods to detect misinformation \cite{sharma2019combating,nasir2021fake,guo2022survey}. In InkSync, verification is open-domain (it requires verifying information not in the document), but limits the scope to individual edits, and puts the author at the center of the process. The system suggests search queries, but the author ultimately decides on searches to conduct and on the ultimate accuracy label.

\section{Survey on Current Text Editing Practices} \label{sec:survey}

\begin{figure*}
    \centering
    \begin{minipage}[t]{0.32\textwidth}
        \subfloat[How often do you edit text documents?]
        {\label{fig:prestudy_plot1} \includegraphics[width=\textwidth]{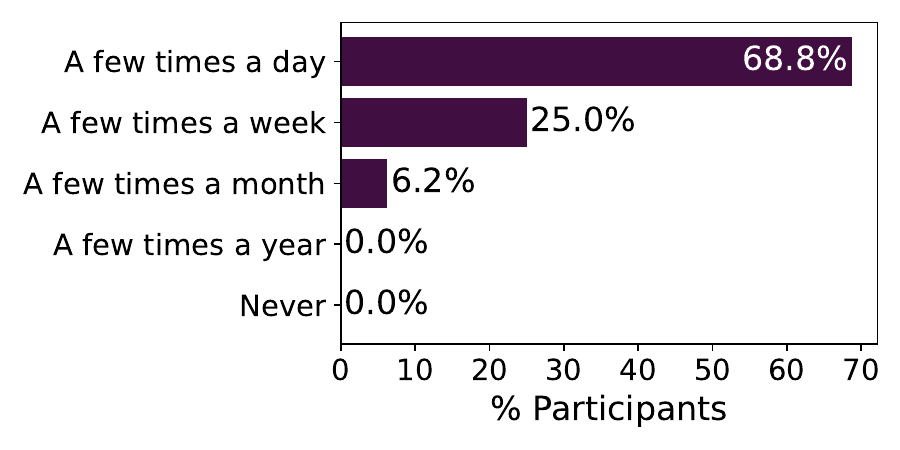}}
    \end{minipage}\hfill
    \begin{minipage}[t]{.32\textwidth}
        \subfloat[What type of documents do you edit?]
        {\label{fig:prestudy_plot2} \includegraphics[width=\textwidth]{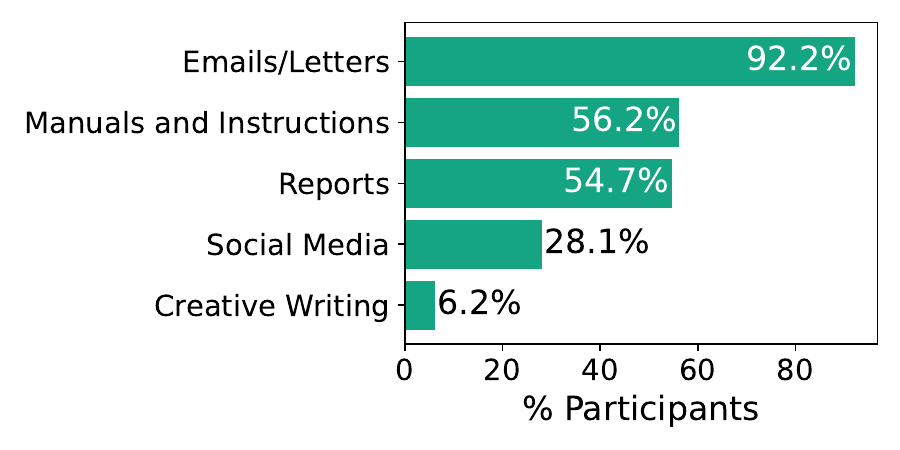}}
    \end{minipage}\hfill
    \begin{minipage}[t]{.32\textwidth}
        \subfloat[How often do you use a chat-based LLM for help with writing?]
        {\label{fig:prestudy_plot3} \includegraphics[width=\textwidth]{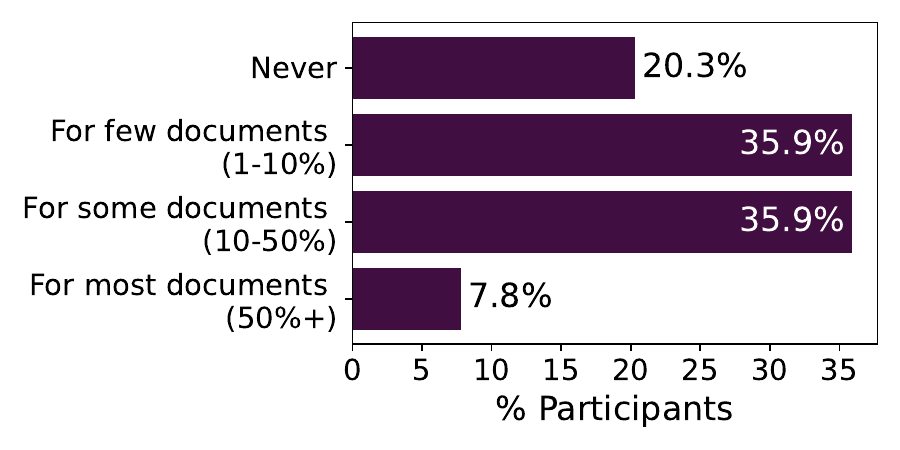}}
    \end{minipage}

    \begin{minipage}[t]{.32\textwidth}
        \subfloat[What type of help do you look for?]
        {\label{fig:prestudy_plot4} \includegraphics[width=\textwidth]{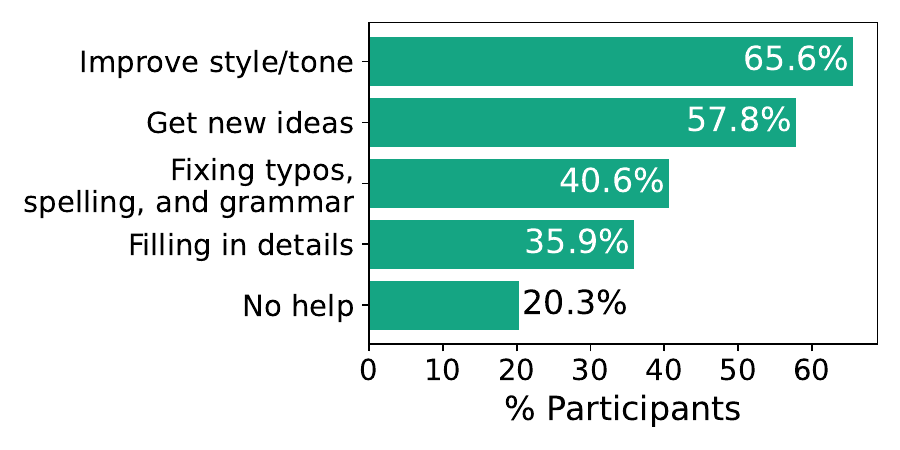}}
    \end{minipage}
    \hfill
    \begin{minipage}[t]{.32\textwidth}
        \subfloat[What do you do when an LLM returns a new version of a document?]
        {\label{fig:prestudy_plot6} \includegraphics[width=\textwidth]{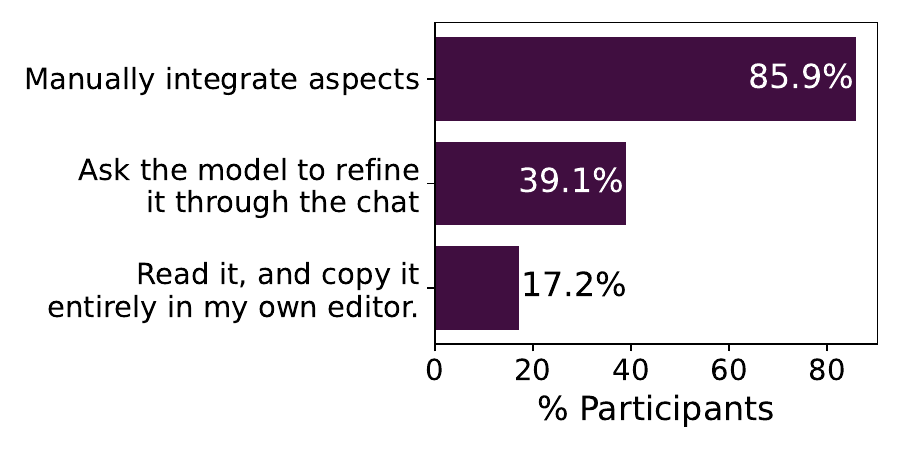}}
    \end{minipage}
    \hfill
    \begin{minipage}[t]{.32\textwidth}
        \subfloat[How do you verify LLM-introduced information?]
        {\label{fig:prestudy_plot5} \includegraphics[width=\textwidth]{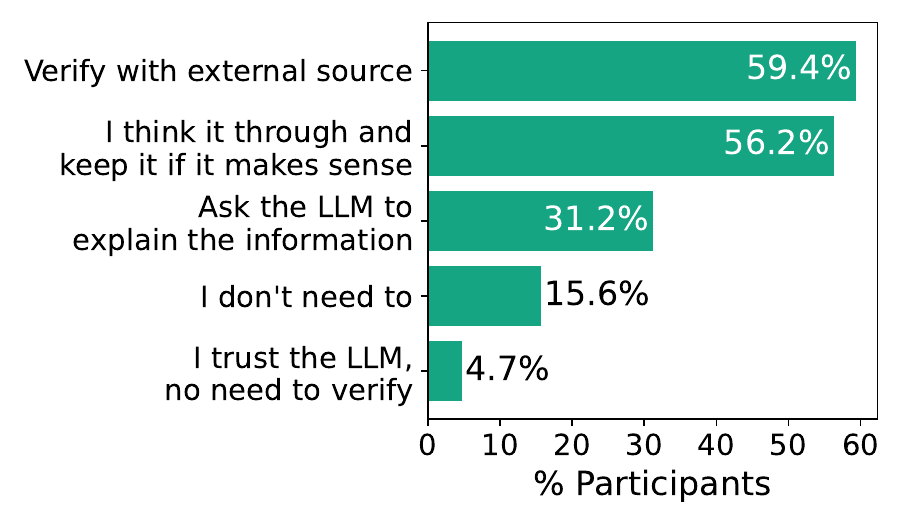}}
    \end{minipage}
    \hfill

    \caption{Responses by 64 surveyed participants on document editing habits and LLM usage in text editing tasks.}
    \label{fig:survey_plots}
\end{figure*}

We conducted a short survey to gain insights into common practices with regard to text editing and the use of LLM-based conversational interfaces to accomplish text editing goals.

\subsection{Survey and Study Participant Recruitment} \label{sec:participant_recruitment}

The first author's organization does not have an IRB approval process in place; instead, the survey and study designs were reviewed by a member of the ethics committee of that organization. Personally identifiable information (email) was collected but not stored, and all participants were informed that anonymized usage data would be collected for the purposes of research. 

The survey was shared publicly on an online messaging platform for US-based knowledge workers and consisted of 6 multiple-choice questions, which we estimate would take 2-3 minutes to complete. Participants were not compensated for completing this survey, however, they were later invited to participate in the usability studies which were of longer duration and were compensated. For the studies, participants were compensated at a rate of 36 USD/hour and were informed that they could leave the study at any time without penalty.

\subsection{Survey Results}

In total 64 participants responded to the survey. We did not ask participants to provide their gender or age but collected participants' roles within the organization: 39\% of participants are engineers (e.g., Software Engineering, Technical Consultant/Architect), 29\% occupy business-related roles (e.g., Product Managers, Accountant), 18\% work in sales-related roles (e.g., Account Executive, Success Manager, etc.), and 14\% occupy technical-writing roles (e.g., Technical Writer, Content Strategist).

Figure~\ref{fig:survey_plots} shows the six survey questions and the aggregated results. Most respondents edit text documents multiple times a day (Figure~\ref{fig:prestudy_plot1}), and most frequently edit emails, manuals, and reports (Figure~\ref{fig:prestudy_plot2}). The following paragraphs summarize the results regarding LLM-based editing.

\paragraph{Figure~\ref{fig:prestudy_plot3}: How often do you use a chat-based LLM for help with writing?} One in five participants responded that they had never used a conversational LLM for such tasks. About two-thirds of the respondents get feedback from an LLM for a few (1-10\%) or some (10-50\%) of the documents they edit, while 8\% are early adopters who use LLMs for most text editing tasks.

\paragraph{Figure~\ref{fig:prestudy_plot4}: What type of help do you look for?} Conversational LLMs are most often used for style and structure guidance (e.g., making an email's tone more professional), and to introduce new information to the document (e.g., brainstorming new ideas). Other uses include fixing grammar and typos and filling in details. With InkSync, we propose several LLM-based components, each targeted at different editing objectives.

\paragraph{Figure~\ref{fig:prestudy_plot6}: What do you do when an LLM returns a new version of a document?} Once a user provides a document in a standard conversational LLMs, the model replies with a revised document, modified to account for the user's query. Yet only 17\% of the participants said that they copied the entirety of the LLM version as a final document. About 39\% of participants reported that they sometimes follow up with the LLM to refine the editing conversationally until they reach a version that fits their needs, while a strong majority of participants report that they manually integrate aspects of the LLM draft back into their draft. Yet manual integration of edits can be onerous on the user: it requires mentally aligning the initial draft with the LLM's proposal, detecting differences, and editing the initial draft to include desirable edits. With InkSync, LLM-based components return executable edits that can be accepted into the working draft or dismissed with one click, increasing transparency and enabling the tracking of author provenance.

\paragraph{Figure~\ref{fig:prestudy_plot5}: How do you verify LLM-introduced information?} Besides a minority who does not report verifying LLM-provided suggestions, most participants are aware that LLM information requires at least to be reviewed, and about 60\% of the respondents rely on an external tool (such as a search engine) to verify information accuracy before integrating it into their draft. In InkSync, the verification process is streamlined by having the system suggest search queries to the user when they request to verify information, which opens in an external search engine.

We note that our findings are representative of a professional population working in the U.S. and may not generalize.

\section{InkSync System Components and Interface} \label{sec:system}

In a standard LLM-based conversational interface, the system replies to a user query (also called a message or a prompt) with an answer formatted in plain text. If the user query includes a document to be edited, the plain text reply may include a revised version of the document along with high-level feedback. The user must then read the plain text reply, and decide which editing suggestions to manually incorporate into the original document.

InkSync conversational components can reply to an author query both with a plain text reply in the chatbox and with a suggested list of \textit{executable edits}. These executable edits are placed \textit{directly within the document} that is being edited and are visually marked via underlines. The underlines remain visible in the document until the author either accepts or dismisses the suggestions. This allows the author to see precisely what the suggestions are and where they occur in the document.
Executable edits require the LLM to generate its edits in a specific format, which can be interpreted by the text editor. 

Figure~\ref{fig:inksync_interface_main} provides an annotated screenshot of the InkSync text editor. The leftmost panel allows for the creation of new documents and the editing of saved documents. The ``Audit Interface'' button in the bottom left allows the author to switch between the edit and the audit view of the active document.
The center main panel displays the latest version of the document being edited in the text editor panel. In the editing panel, the author can manually edit the document, or view, accept, and dismiss system suggestions. The rightmost panel displays the editing controls, including the Markers, Chat, Comment, and Verify tabs. Additional interface features are shown in Figure~\ref{fig:inksync_additional_components}.

In this section, we first explain the system components of InkSync  (Section~\ref{sec:system_details}). Next, we describe the interface components in detail: the four edit-suggesting components (Section~\ref{sec:edit_suggesting_components}), and the Warn-Verify-Audit framework (Section~\ref{sec:warn-verify-audit}).


\subsection{InkSync System Components}
\label{sec:system_details}
\subsubsection{Executable Editing Language} \label{sec:executable_edit_language}

In InkSync, when an LLM-based component suggests executable edits to the author, it must generate the edits in a JSON format to be used in the editing tool. This representation includes the original text, the suggested replacement text, and a flag indicating if the replacement text includes new information. The schema for the JSON is shown in this example:
\begin{myjsonblock}{Executable Edit Language - Example Edit}
    \begin{minted}[fontsize=\small]{json}
{
 "original_text": "Lets plan a trip too Paris.",
 "replace_text":  "Let's plan a trip to Paris.",
 "component":     "marker_TYPO",
 "replace_all":   "0",
 "new_info":      "0"
}
    \end{minted}
\end{myjsonblock}

The definitions for the keys and acceptable values are the following:
\begin{itemize}
    \item \texttt{original\_text}: A string exactly present in the draft. If the string is not present at least once in the draft, the edit is discarded,
    \item \texttt{replace\_text}: A string to replace the \texttt{original\_text} string with when the edit is accepted by the user,
    \item \texttt{component}: The component that originated the edit (options \texttt{marker}, \texttt{chat}, \texttt{comment}, and \texttt{brainstorm}, and an optional subcomponent (e.g., \texttt{marker\_typo}),
    \item \texttt{replace\_all} (optional): If the value is set to \texttt{1}, and the \texttt{original\_text} string appears more than once in the draft, multiple edit suggestions are shown to the user, otherwise, only the first occurrence is suggested as an edit. Default value is \texttt{0},
    \item \texttt{new\_info} (optional): If the value is set to \texttt{1}, the edit is marked as introducing new information to the document. The default value is \texttt{0}.
\end{itemize}

\begin{figure*}
    \centering
    \includegraphics[width=0.98\textwidth]{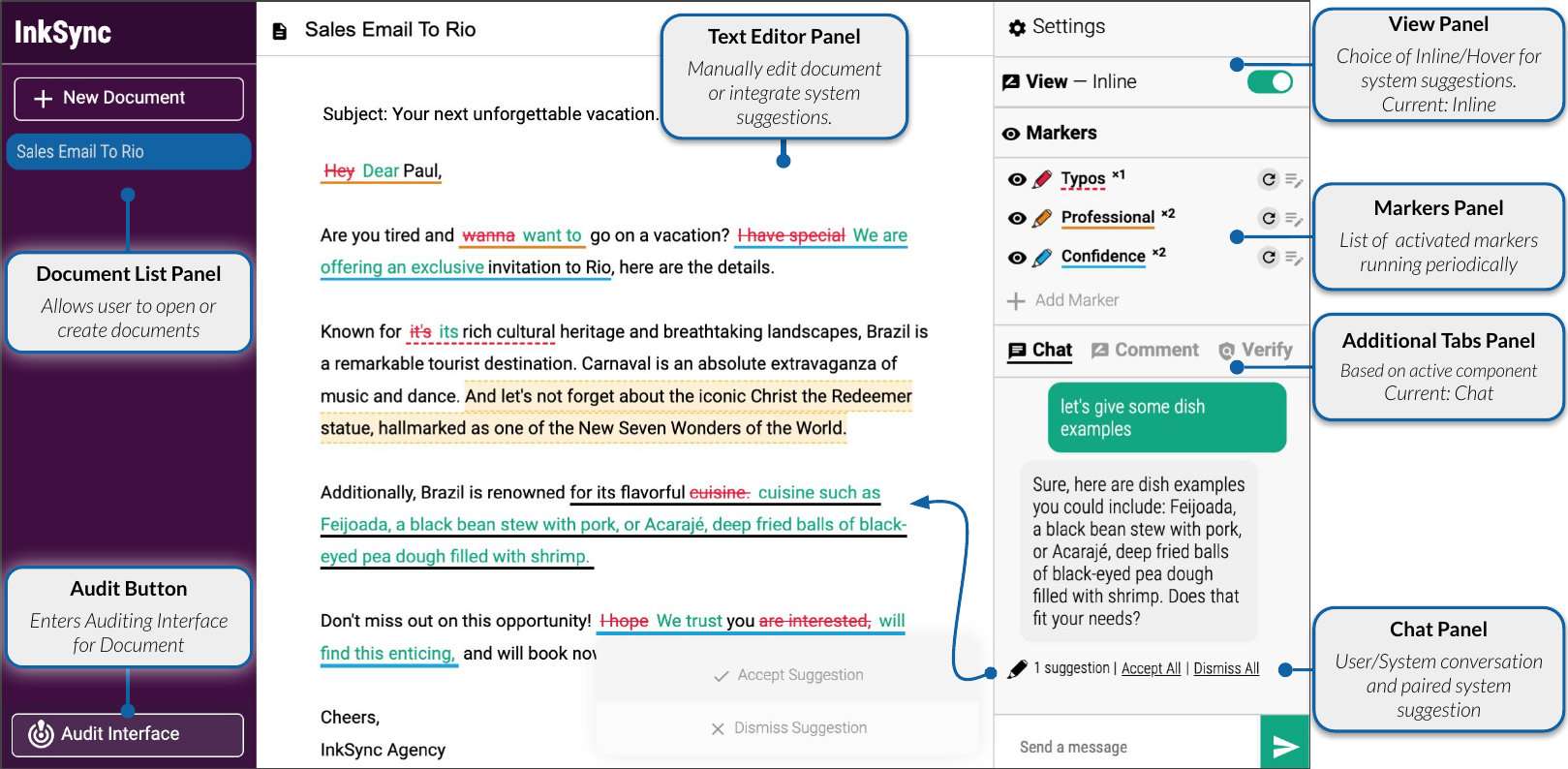}
    \caption{The InkSync text editing interface layout. \textbf{Left Panel:} facility to create new, edit, or audit existing documents. \textbf{Center Panel:} The editing panel shows the current document version in black and edit suggestions in red strikeout and green new text. Colored underlines show the source of the suggested edits. Yellow highlighted background indicates a text span that has been highlighted with the Comment feature. Clicking on the yellow text activates the Comment tab in the lower right panel. An Accept/Dismiss dialogue menu is shown beneath the lowermost suggestion. \textbf{Right Panel:} contains customizable settings, the Markers panel, as well as Chat, Command, and Verify components. In the figure, the author has queried the Chat component with ``let's give some dish examples'' and the edit suggestion is shown in the editing panel with a black underline. The curved arrow shows the location of the suggested edit and is not part of the interface.}
    \label{fig:inksync_interface_main}
\end{figure*}

\subsubsection{Representing Edit Suggestions} \label{sec:rep_edit_suggestions}

In the interface, each executable edit suggestion\footnote{In this paper we also refer to an executable edit suggestion as an edit suggestion or an edit.} is underlined using a style (e.g., solid, dotted, etc.) and color that identifies the component of the system that created the edit. The author can hover over an edit suggestion to view an overlaid menu to Accept or Dismiss the edit.

When presenting an edit suggestion in the interface, a word-level sequence alignment is computed between the \texttt{original\_text} and \texttt{replace\_text} strings using the Levenshtein algorithm \cite{Levenshtein1966BinaryCC}, and displayed with colors indicated \textcolor{colorins}{inserted} and \textcolor{colordel}{deleted} words. The alignment sequence and edit alignment visualization for the example edit is:

\begin{myjsonblock}{Edit Alignment Visualization}
    \textcolor{colordel}{\sout{Lets}} \textcolor{colorins}{Let's} plan a trip \textcolor{colordel}{\sout{too}} \textcolor{colorins}{to} Paris.
\end{myjsonblock}

When an edit is accepted, the \texttt{original\_text} is replaced by the \texttt{replace\_text}. When an edit is dismissed, the \texttt{original\_text} remains unchanged. In both cases the editing interface is ``cleaned up'': the edit's underlines and strike-through are removed, visually clearing the edit from the text editor.

The author can choose between two view modes for the edit suggestions: \texttt{Inline} and \texttt{Hover}. In Inline mode, the edit alignment visualization is directly embedded in the text editor panel. In Hover mode, only the \texttt{original\_text} is shown in the text editor panel, and the alignment sequence is shown in an overlaid window.
The edit view mode can be controlled from the View Panel, in the right-most Settings column (illustrated in Figure~\ref{fig:inksync_additional_components}).

When designing this editing interface, we had to make choices among a number of tradeoffs. Our objective is to maximize edit suggestion clarity, and with this goal in mind we made the following decisions:

\paragraph{Design Choice 1: No Overlapping Suggestions.} In some cases, components might suggest more than one possible edit for the same span of text. In cases where multiple edit suggestions overlap in the \texttt{original\_text}, we choose to display only the most recent edit suggestion.

\paragraph{Design Choice 2: Contextualized Edits.} A choice must be made about how much context to provide around an edit suggestion. We chose to provide context to aid interpretation. For instance, when fixing a typo ``a trip too Paris'', the system should favor ``trip \textcolor{colordel}{\sout{too}} \textcolor{colorins}{to} Paris'' over ``\textcolor{colordel}{\sout{too}} \textcolor{colorins}{to}''.

\paragraph{Design Choice 3: Granular Edits.} We chose to have the system avoid suggesting a single edit that majorly rewrites a paragraph or the entire document. Instead, we favor \textit{granular} edits, which give the author more fine-grained control by breaking down the system suggestions into smaller editing units.

\paragraph{Design Choice 4: Client-Side is Source of Truth.} The author has the ability to edit the document continuously, including while any of the system's components are simultaneously generating suggestions. This can lead to system suggestions being outdated or invalid. In such cases, the client-side version of the document is considered the source of truth, and system suggestions that are no longer executable (i.e., their \texttt{original\_text} is no longer present in the document due to author edits) are marked in the document's revision history as \textit{implicitly dismissed}.

\subsubsection{Prompt Design} \label{sec:prompt_design}

Each of the four edit-suggesting components is implemented with a single prompt to an LLM. Each prompt is composed of (1) a high-level description of the component, (2) the latest version of the document, (3) author queries when applicable, and (4) the design choices of the InkSync system. Each prompt further provides 3 examples on simplified single-sentence documents to illustrate desired behavior and output format. Appendix~\ref{app:llm_chat_prompt} concretely illustrates the prompt design used for the Chat component.

For all prompts, the LLM is expected to return a valid JSON string that follows a predefined schema and contains the list of executable edit suggestions from the system. In cases where the LLM does not return a valid JSON, a system error is forwarded to the user, suggesting to try again. 

The InkSync system is compatible with any LLM since each component's prompt defines the information format it requires to operate. In the two usability studies we conducted, we used OpenAI's GPT-4 LLM \cite{OpenAI2023GPT4TR}, as we found in initial experiments that it had a lower rate of failure due to invalid output formatting (roughly 2-5\%), and produced higher quality edit suggestions than smaller alternatives such as GPT3.5-Turbo. GPT-4's size leads to slower system responses, and we analyze system response time in Appendix~\ref{app:study1_response_time}.

\subsection{Edit-suggesting Components} \label{sec:edit_suggesting_components}

\begin{figure*}
    \centering
    \includegraphics[width=0.98\textwidth]{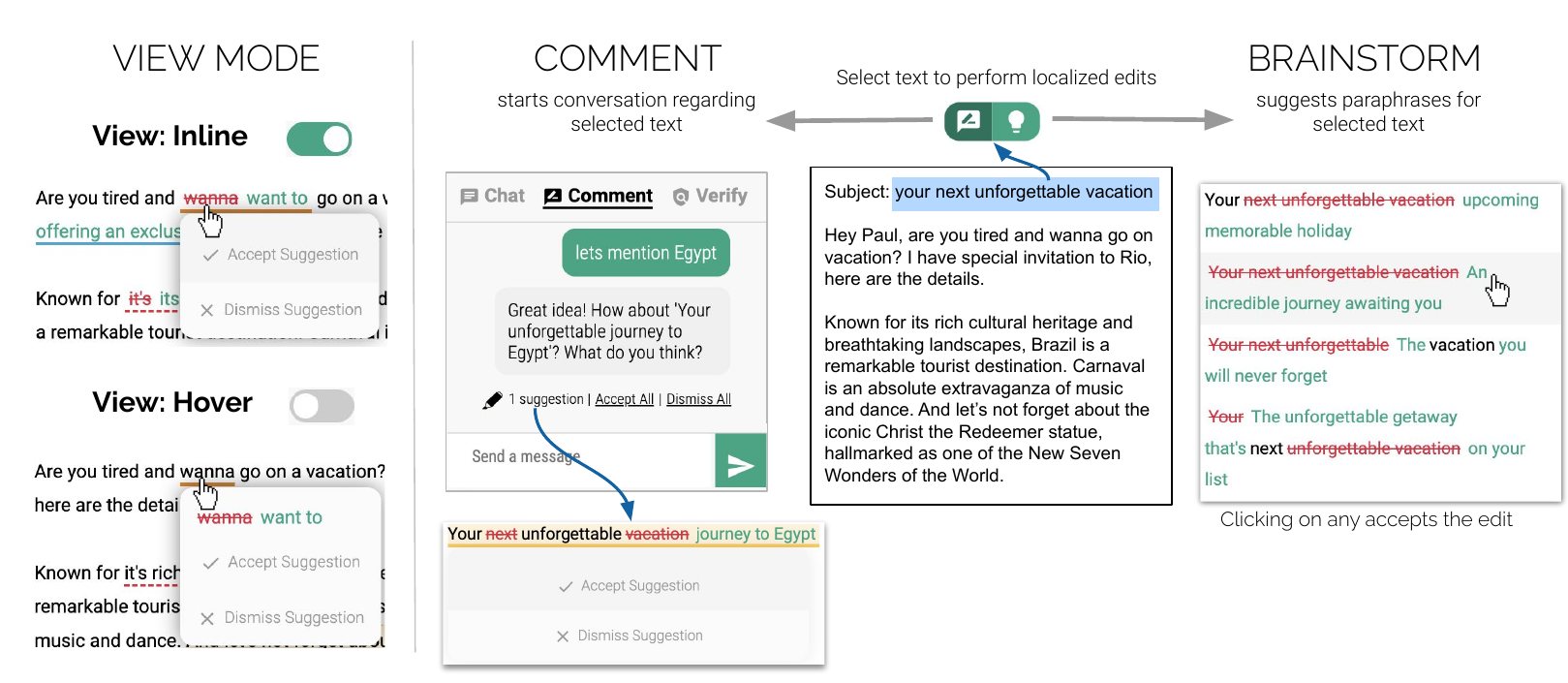}
    \caption{Additional features of the InkSync edit interface: (left) edit view modes for edit suggestions, (right) Comment and Brainstorm component workflows for localized edits.}
    \label{fig:inksync_additional_components}
\end{figure*}
We now present the four components that can suggest edits in InkSync: Chat, Comment, Markers, and Brainstorm.

\subsubsection{\symbolimg{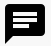} Chat} \label{sec:chat}

A Chat panel resides in the bottom-right portion of the InkSync interface. The author can send a message to the chat when seeking information (e.g., ``What is the capital of Argentina?'') or requesting edit suggestions (e.g., ``Locate and fix all typos'').

When the author sends a message, the system replies with (1) a plain text \textit{reply} shown in the Chat panel, and (2) some number of suggested edits that are shown in the main text editing panel (in our studies, the number of edit suggestions in response to a query ranged from 0 to 14). In Figure~\ref{fig:inksync_interface_main}, the user prompted the system with: ``let's give some dish examples'' and the system responded with three dish suggestions which are displayed both as an executable edit and summarized in the conversational reply panel in the lower right. In cases where the user is seeking information (e.g., ``what is the capital of Argentina?'') and not editing suggestions, the system is prompted to return solely a textual reply.

Importantly, the system autonomously determines the number of edits to suggest to the author.
Chat-originated edits are integrated into the main text editor panel and indicated with a solid black line. When there are multiple suggested chat-originated edits, the bottom of the chat panel indicates the number of edits, and gives the option to ``Accept All'' or ``Dismiss All''. This enables the user to accept chat edits individually (in the main panel) or as a group (in the Chat panel).

\subsubsection{\symbolimg{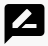} Comment} \label{sec:comment}

The Chat component makes suggestions applicable to the entire document but does not allow the author to efficiently specify the desired location for edits.
The Comments component allows more fine-grained control in a conversational-style interface. 
For example, rather than querying the global chat with: ``Let's add more details at the end of the fifth paragraph'', the author can select a few words at the end of the fifth paragraph, create a Comment, and query: ``Let's add more details''.

Comments work in three steps, illustrated in Figure~\ref{fig:inksync_additional_components}. The author first selects a span of text in the main editing panel, which causes a small dialogue box to pop up over the selected text. If the author clicks on the \symbolimg[0.3cm]{figures/icons/comment.png} Comment icon that appears below the selection, this causes a Comment to be created, signaled by a \hlcomment{yellow background fill}.

Next, clicking on the yellow Comment span opens a localized, comment-specific conversation in the Additional Tabs panel. When the author enters a query into this panel, the system is provided with the text from the selected span and is instructed to focus its suggestions on that text and the nearby context.
As with other edit suggestions, the author can choose to accept or dismiss the suggestion. Finally, the author can click ``Resolve'' in the Comment panel which dismisses pending Comment suggestions and hides the yellow comment span from the edit view.

\subsubsection{\symbolimg{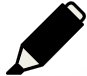} Markers} \label{sec:markers}

The Chat and Comment components are both reactive and only suggest edits upon user request. The Markers component on the other hand is a collection of proactive edit-suggesting tools that run periodically in the background during an editing session and insert one or more edit suggestions directly into the document. 

Each marker is designed to suggest edits for one specialty (e.g., fixing typos, formal tone writing, etc.), and has a corresponding name (Typo, Formal, Emojis, etc.) and underline style and color. Figure~\ref{fig:inksync_interface_main} shows marker suggestion examples: the Professional marker suggests an edit to replace ``Hey Paul'' with ``Dear Paul'' with a solid orange underline.

The Markers Panel in the interface lists active markers and allows the author to control their behavior. The author can toggle each marker to make all of its suggestions \symbolimg[0.3cm]{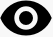} visible or \symbolimg[0.3cm]{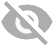} hidden, and can trigger an individual marker to rerun with the refresh icon (\symbolimg[0.3cm]{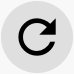}). Finally, a marker can be edited and deleted through a marker editing menu (\symbolimg[0.3cm]{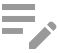}).

The Markers component functions by providing the exact list of active markers and their descriptions in the prompt of the LLM that suggests edits. This gives the author the option to create a new marker, by choosing a marker name,  an underlining color and style, and an optional description that can provide precise instructions and examples. 

\subsubsection{\symbolimg{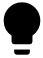} Brainstorm} \label{sec:brainstorm}

We made a design choice of disallowing more than one edit suggestion for the same span of text. This choice can be limiting when an author is looking for multiple rephrasing options for a given passage (e.g., help brainstorm options for the headline of a news article).

The brainstorm component \symbolimg[0.3cm]{figures/icons/brainstorm.png} allows authors to tell the system to suggest multiple ideas for a given span of text. 
A brainstorm is initiated by the author in the same way as the Comment, by selecting a passage of text in the document and viewing the Comment/Brainstorm dialogue box. Once the author clicks on the brainstorm icon, the system is instructed to suggest 3-5 diverse paraphrases of the selected text, which are shown in a drop-down menu just beneath the selected text. The author can then accept any of the provided options, or dismiss the brainstorm window (see Figure~\ref{fig:inksync_additional_components} for an example Brainstorm interaction).

\subsubsection{\symbolimg{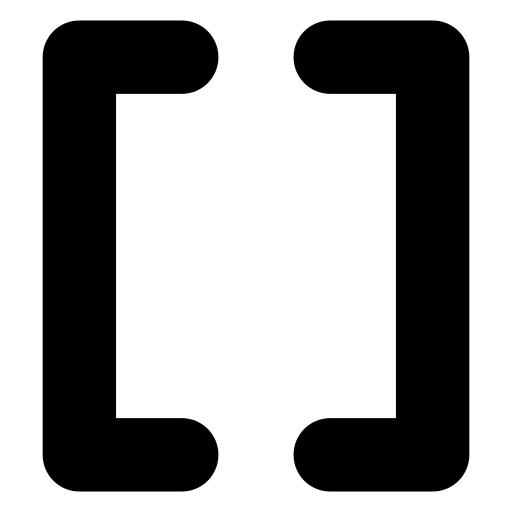} Bracket Shortcuts} \label{sec:shortcuts}

The four edit-suggesting components require the user to leave the text editing panel for activation, through clicking (Markers, Brainstorm), or writing in a separate dialogue box (Chat, Comment), which might disrupt the user's focus. We implement bracket shortcuts that allow the author to initiate a Comment or Brainstorm from the text editor.

The user can put any passage within [square brackets], which the system initially classifies as a \textit{command} or \textit{content}. A command is any extraneous text that will not be part of the final document (e.g., [add more detail here]), whereas content is a passage that is an integral part of the document (e.g., Egypt is a [very pretty] place).

If the system determines that the bracketed text is a command, it initiates a comment and automatically sends the command as the first user message in the conversation. If the bracketed text is categorized as content, a brainstorm is initiated. In short, the bracket shortcut lowers the effort required to initiate a Comment or Brainstorm.

\subsection{Warn-Verify-Audit Framework} \label{sec:warn-verify-audit}

\begin{figure*}
    \centering
    \includegraphics[width=0.9\textwidth]{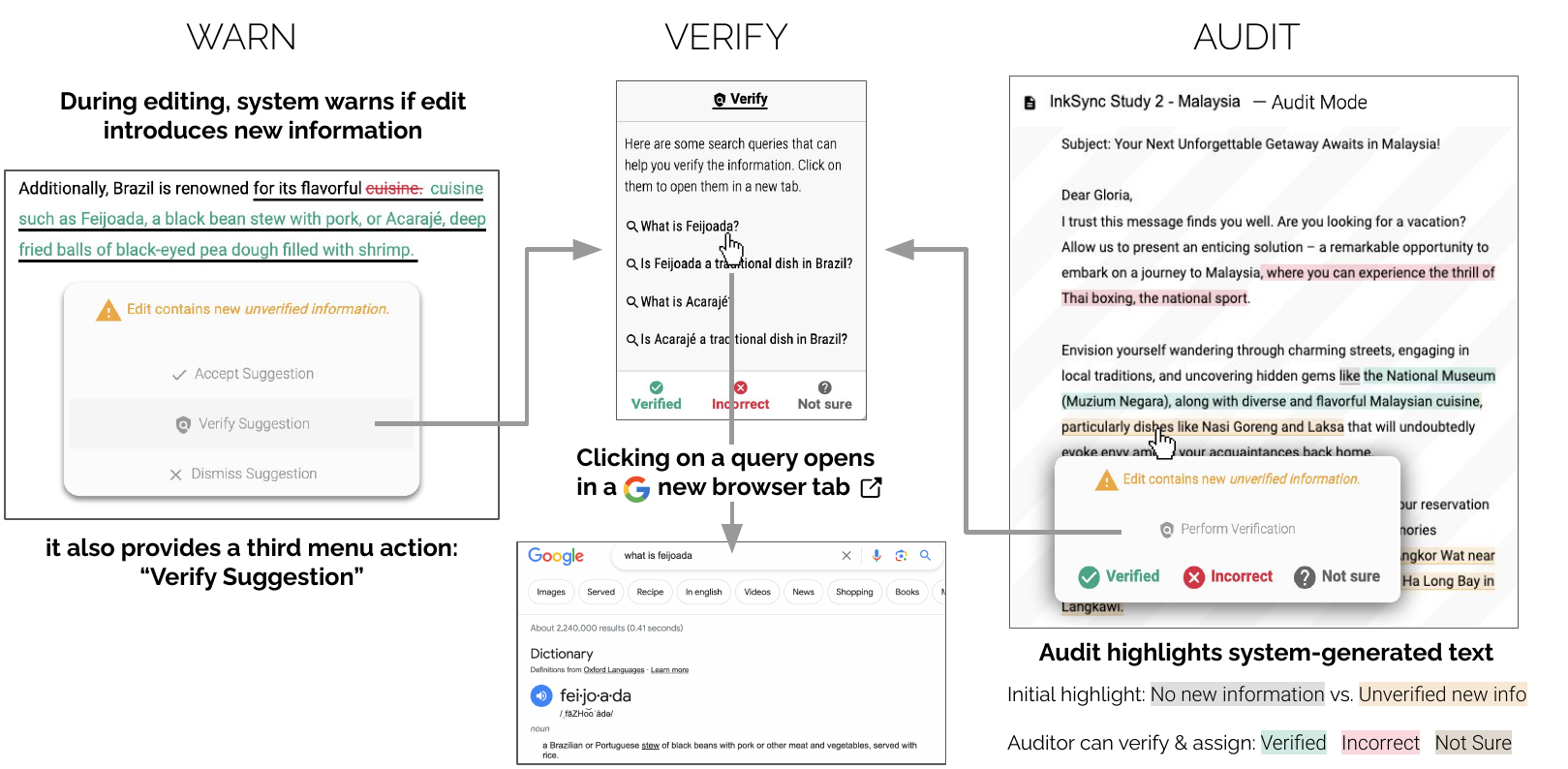}
    \caption{Overview of the Warn, and Verify components and Audit interface in the InkSync system. During editing (left), when a suggested edit introduces new information, the system displays a visual warning and provides an option to Verify the suggestion. During auditing, the system traces and highlights all system-generated content, and also provides an opportunity to verify the traced content. When the author/auditor initiates a verification, the system generates a list of search engine queries, which open in a new browser tab when clicked on.}
    \label{fig:warn_verify_audit}
\end{figure*}

\subsubsection{\symbolimg[0.3cm]{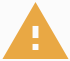} Warn} \label{sec:warn}

Some of the suggested edits might introduce new information to the document, which might be factually incorrect. Factual inaccuracies in system responses can both result in the highly undesirable outcome of introducing incorrect information into communication and can also negatively affect users' trust in LLM-based systems. Therefore, we introduce a fact verification aid into InkSync.

Our verification pipeline relies on the observation that even though the LLM might not be capable of detecting if newly introduced information is inaccurate, it might be capable of assessing if an edit introduces new information that goes beyond what was written by the author in the document. Having a notice about new information could be used to warn the author and give them an opportunity to verify the information before accepting the edit.
Therefore, the InkSync prompt is designed such that the LLM augments all edit-suggesting components of InkSync to assign a binary label to the edits they generate, indicating whether the edit introduces new information or not (the \texttt{new\_info} key in the executable edit language).

We performed a manual analysis to determine how accurate the detection of new information is. We randomly selected 200 edits from the 12,000+ edits the system suggested to participants in the usability studies, selecting 100 edits the system predicted as \texttt{new\_info=0}, and 100 with \texttt{new\_info=1}. We shuffled the samples, hid the system label predictions, and performed a manual annotation of the label.

We find that the manual label agrees with the system-predicted \texttt{new\_info} prediction 97.5\% of the time. Inspection of the five disagreement cases reveals that they are ambiguous borderline cases (e.g., ``Your \textcolor{colordel}{dream vacation} \textcolor{colorins}{gastronomic adventure} awaits!''). This small-scale manual annotation confirms empirically that GPT-4 can predict when an edit introduces new information accurately.

As shown in Figure~\ref{fig:warn_verify_audit} (left), when the system detects that a suggested edit introduces new information, it augments the Accept/Dismiss menu to show the author a visual warning which states: ``\symbolimg[0.3cm]{figures/icons/warn.png} Edit contains new \textit{unverified information}''.

\subsubsection{\symbolimg{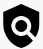} Verify} \label{sec:verify} 

In addition to the warning, a ``\symbolimg[0.3cm]{figures/icons/verify.png} Verify'' action is inserted into the menu. If the user clicks on the Verify button, the Verify component prompts an LLM to generate a list of search engine queries that can be used to verify the new information in the edit. The search engine queries are presented in the interface in a verification panel (Figure~\ref{fig:warn_verify_audit}, center). If the user clicks on a generated search engine query, the system opens a new browser tab to the Google search engine populated with the query.

Automatically generating search engine queries is intended to reduce the amount of work required from the author to verify system-introduced information. Opening the search engine query in a separate interface allows the user to refine and deepen the search before coming back to the InkSync interface.

Once the author forms an opinion on the accuracy of the edit, they can optionally mark the edit with one of three labels: ``Verified'', ``Incorrect'', and ``Not Sure'', and can make an informed decision on whether to accept the edit into their document or dismiss it. The Incorrect mark can be used to remind the author that they want to come back later and change that piece of information.

\subsubsection{\symbolimg{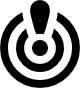} Auditing Interface} \label{sec:auditing_interface}

In the editing interface, once an author accepts an edit suggestion, the changes are stored and no longer visible. As the editing session progresses, the author might lose track of the edit suggestions' content they accepted.

To facilitate provenance tracing within the document, we developed an Audit interface that complements the Edit interface. The Audit interface empowers an auditor -- optionally distinct from the author -- to efficiently review the product of the editing process, confirm the accuracy of system-generated content, and potentially flag inaccuracies. The Audit Interface (Figure~\ref{fig:warn_verify_audit}) provides a non-editable view of the document, highlighting any system-generated content present in the document from previously accepted edits during editing.

In the Audit interface, any system-originated content is highlighted using a color scheme based on whether it introduces new information (yellow) or not (grey). If the auditor hovers over system-generated content, they are shown an action menu that allows them to Verify the edit (process described in Section~\ref{sec:verify}) or directly assign a verification label. Once the auditor assigns a verification label, its highlighting color is modified to green (Verified), red (Incorrect), or Orange (Not Sure).

If the mode is switched from Audit to Edit, the author can resume editing the draft, and the verification information is made visible in that mode. The tracing of system-generated content requires a character-level alignment algorithm, which we detail in Appendix~\ref{app:tracing_algorithm}.

\section{Usability Study 1: Interaction Style Evaluation} \label{sec:study1}

We conducted a 20-minute within-participants usability study with 55 participants, framed around the task of customizing sales emails for a fictional InkSync travel agency. The objectives were to:
\begin{enumerate}
    \item Assess the usability of the four edit-suggesting components of InkSync, in comparison to existing baselines,
    \item Determine which editing objectives each InkSync component is most useful for,
    \item Ascertain participant preferences about the conditions included in the study.
\end{enumerate}

\subsection{Participants} \label{sec:study1_participants}

We recruited participants from our initial survey respondents and added referred participants to the study by initial participants. Thus, Study 1 participants do not entirely overlap with initial survey respondents.

We reviewed the tasks completed by all participants and found that all participants understood the task clearly and applied themselves enough such that no participant filtering was required. Three participants completed the study during an OpenAI outage, which caused the system to have limited operability. Those participants were paid, but their data was excluded, resulting in 52 participants' data being included in the results and analysis.

\subsection{Study Procedure} \label{sec:study1_procedure}

The study consisted of three 6-minute editing sessions and a 2-minute survey, for a total of 20 minutes. The study followed a partial within-participants design, in that each participant completed a task with 3 out of the 6 interface conditions. Therefore, each interface condition was seen by 29-35 participants. We randomized the selection and order of interface conditions and assigned tasks.

\subsubsection{Study Task: Email Customization}

Each session began with a one-minute introductory slide deck on the current condition's features, followed by a five-minute task to edit and customize an email template from a fictional travel agency. Each task provides (1) an initial email template, (2) a fictional recipient name, (3) a target travel destination, and (4) a persona trait for the participant. Appendix~\ref{app:study_customization_options} provides the exact destination and persona traits used. 

The participants were provided two editing objectives: first, to ensure the email had a formal tone and was professional, and second, to customize the email by providing specific suggestions relevant to the recipient's persona and travel destination. In the introductory material, participants were shown an example customized email for a persona and target destination that was not included in the study conditions (a lover of the Opera going to Rome).

Since each participant completed three tasks, we prepared three generic email templates each containing several informal phrases (e.g. ``wanna go on a vacation''), and grammatical errors (e.g., ``Dont miss out''), randomly assigned. The original templates ranged from 100 to 250 words in length. Appendix~\ref{app:study_customization_options} shows one of the email templates used and Figure~\ref{fig:study1_email_example} in the Appendix provides a concrete example of a study participant's editing session.

\subsubsection{Six Interface Conditions}

We included six interface conditions in the study:

\paragraph{Manual Editing.} In this baseline condition, none of the edit-suggesting components are enabled. Participants manually edited the email template, mimicking the process of editing an email in a basic text editor without any assistance.

\paragraph{Non-Executable (NoEx) Chat.} In this baseline condition, the participant has access to the Chat component, but the prompt used for the model is modified, such that the LLM is instructed to only provide its reply in the conversation dialog box, and cannot suggest executable edits. This condition simulates an author using a standard conversational LLM interface (e.g., ChatGPT opened in a different browser tab).

\paragraph{Markers Only.} In this condition, the participant can only interact with the Markers component. Three initial markers are added by default (Typos, Professional, Formal). This condition simulates an author having access to a text editor with a modern text checker.

\paragraph{Chat Only.} In this condition, the participant can interact with the complete InkSync Chat component, which responds to author queries with a plain text reply and executable edits placed into the document that is being edited.

\paragraph{Localized Only.} In this condition, only the two components that require explicit localization from the user -- Comment and Brainstorm -- are enabled in the interface.

\paragraph{4-Comp} In this condition, the participant has access to all four edit-suggesting components of InkSync: Chat, Comment, Markers, and Brainstorm.

Note that none of the conditions contained the Warn, Verify, or Audit components of InkSync since those components are evaluated in a follow-up study, Usability Study 2 (Section~\ref{sec:study2}).

Since participants completed the study unmoderated, we implemented additional controls in the interface -- details in Appendix~\ref{sec:study1_quality_control} -- to help ensure the validity of the findings.

\subsection{Evaluation Methodology and Results} \label{sec:study1_evaluation}

\begin{table}[]
    \centering
    \resizebox{0.95\textwidth}{!}{%
    \begin{tabular}{lcccccc}
    & \multicolumn{6}{c}{\textbf{Interface Condition}} \\
    \cmidrule{2-7}
    Metric / Question & Manual & NoEx Chat & Markers & Localized & Chat & 4-Comp \\
    \midrule
    \texttt{Objective A}: \# Typos, Informal Phrases ($\downarrow$) & \cellcolor[rgb]{0.99, 1.00, 1.00}12.9 \footnotesize{$\pm$0.8} & \cellcolor[rgb]{0.99, 0.99, 0.99}12.8 \footnotesize{$\pm$0.7} & \cellcolor[rgb]{0.57, 0.83, 0.77}7.0 \footnotesize{$\pm$0.7} & \cellcolor[rgb]{0.93, 0.97, 0.96}12.0 \footnotesize{$\pm$0.8} & \cellcolor[rgb]{0.90, 0.96, 0.95}11.6 \footnotesize{$\pm$0.9} & \cellcolor[rgb]{0.44, 0.78, 0.70}5.2 \footnotesize{$\pm$0.8} \\

    \texttt{Objective B}: \# Custom Recommendations ($\uparrow$) &\cellcolor[rgb]{0.80, 0.92, 0.89} 2.9 \footnotesize{$\pm$0.5} &\cellcolor[rgb]{0.60, 0.84, 0.79} 4.3 \footnotesize{$\pm$0.7} &\cellcolor[rgb]{1.00, 1.00, 1.00} 1.5 \footnotesize{$\pm$0.3} &\cellcolor[rgb]{0.48, 0.80, 0.73} 5.1 \footnotesize{$\pm$0.6} &\cellcolor[rgb]{0.43, 0.78, 0.70} 5.5 \footnotesize{$\pm$0.8} &\cellcolor[rgb]{0.51, 0.81, 0.74} 4.9 \footnotesize{$\pm$0.6} \\
    \midrule
    I could complete the task in the provided time. &\cellcolor[rgb]{0.94, 0.97, 0.97} 3.6 \footnotesize{$\pm$0.2} &\cellcolor[rgb]{0.87, 0.95, 0.93} 3.7 \footnotesize{$\pm$0.2} &\cellcolor[rgb]{0.68, 0.87, 0.83} 4.0 \footnotesize{$\pm$0.2} &\cellcolor[rgb]{0.81, 0.92, 0.90} 3.8 \footnotesize{$\pm$0.2} &\cellcolor[rgb]{0.68, 0.87, 0.83} 4.0 \footnotesize{$\pm$0.2} &\cellcolor[rgb]{0.61, 0.85, 0.79} 4.1 \footnotesize{$\pm$0.2} \\
    I felt in control of the editing process. &\cellcolor[rgb]{0.48, 0.80, 0.73} 4.3 \footnotesize{$\pm$0.1} &\cellcolor[rgb]{0.74, 0.90, 0.86} 3.9 \footnotesize{$\pm$0.2} &\cellcolor[rgb]{0.61, 0.85, 0.79} 4.1 \footnotesize{$\pm$0.2} &\cellcolor[rgb]{0.68, 0.87, 0.83} 4.0 \footnotesize{$\pm$0.2} &\cellcolor[rgb]{0.48, 0.80, 0.73} 4.3 \footnotesize{$\pm$0.2} &\cellcolor[rgb]{0.61, 0.85, 0.79} 4.1 \footnotesize{$\pm$0.1} \\
    I can easily adjust system suggestions. & - &\cellcolor[rgb]{0.94, 0.97, 0.97} 3.6 \footnotesize{$\pm$0.2} &\cellcolor[rgb]{0.81, 0.92, 0.90} 3.8 \footnotesize{$\pm$0.2} &\cellcolor[rgb]{0.81, 0.92, 0.90} 3.8 \footnotesize{$\pm$0.2} &\cellcolor[rgb]{0.74, 0.90, 0.86} 3.9 \footnotesize{$\pm$0.2} &\cellcolor[rgb]{0.55, 0.82, 0.76} 4.2 \footnotesize{$\pm$0.1} \\
    System suggestions help complete the task. & - &\cellcolor[rgb]{0.61, 0.85, 0.79} 4.1 \footnotesize{$\pm$0.1} &\cellcolor[rgb]{0.61, 0.85, 0.79} 4.1 \footnotesize{$\pm$0.2} &\cellcolor[rgb]{0.61, 0.85, 0.79} 4.1 \footnotesize{$\pm$0.2} &\cellcolor[rgb]{0.42, 0.77, 0.69} 4.4 \footnotesize{$\pm$0.1} &\cellcolor[rgb]{0.48, 0.80, 0.73} 4.3 \footnotesize{$\pm$0.1} \\
    System suggestions are easy to understand. & - &\cellcolor[rgb]{0.68, 0.87, 0.83} 4.0 \footnotesize{$\pm$0.2} &\cellcolor[rgb]{0.68, 0.87, 0.83} 4.0 \footnotesize{$\pm$0.2} &\cellcolor[rgb]{0.68, 0.87, 0.83} 4.0 \footnotesize{$\pm$0.1} &\cellcolor[rgb]{0.42, 0.77, 0.69} 4.4 \footnotesize{$\pm$0.1} &\cellcolor[rgb]{0.42, 0.77, 0.69} 4.4 \footnotesize{$\pm$0.1} \\
    System suggestions are easy to integrate. & - &\cellcolor[rgb]{1.00, 1.00, 1.00} 3.2 \footnotesize{$\pm$0.2} &\cellcolor[rgb]{0.61, 0.85, 0.79} 4.1 \footnotesize{$\pm$0.2} &\cellcolor[rgb]{0.68, 0.87, 0.83} 4.0 \footnotesize{$\pm$0.2} &\cellcolor[rgb]{0.48, 0.80, 0.73} 4.3 \footnotesize{$\pm$0.1} &\cellcolor[rgb]{0.55, 0.82, 0.76} 4.2 \footnotesize{$\pm$0.1} \\
    Specifying the location for desired edits is simple. & - &\cellcolor[rgb]{1.00, 1.00, 1.00} 3.4 \footnotesize{$\pm$0.2} &\cellcolor[rgb]{0.87, 0.95, 0.93} 3.7 \footnotesize{$\pm$0.2} &\cellcolor[rgb]{0.68, 0.87, 0.83} 4.0 \footnotesize{$\pm$0.2} &\cellcolor[rgb]{0.68, 0.87, 0.83} 4.0 \footnotesize{$\pm$0.2} &\cellcolor[rgb]{0.61, 0.85, 0.79} 4.1 \footnotesize{$\pm$0.2} \\
    The system is reliable and responds as expected. &\cellcolor[rgb]{1.00, 1.00, 1.00} 3.5 \footnotesize{$\pm$0.2} &\cellcolor[rgb]{0.87, 0.95, 0.93} 3.7 \footnotesize{$\pm$0.2} &\cellcolor[rgb]{0.81, 0.92, 0.90} 3.8 \footnotesize{$\pm$0.2} &\cellcolor[rgb]{0.87, 0.95, 0.93} 3.7 \footnotesize{$\pm$0.2} &\cellcolor[rgb]{0.61, 0.85, 0.79} 4.1 \footnotesize{$\pm$0.2} &\cellcolor[rgb]{0.61, 0.85, 0.79} 4.1 \footnotesize{$\pm$0.1} \\
    The system responds in a timely manner. &\cellcolor[rgb]{1.00, 1.00, 1.00} 3.5 \footnotesize{$\pm$0.2} &\cellcolor[rgb]{0.55, 0.82, 0.76} 4.2 \footnotesize{$\pm$0.2} &\cellcolor[rgb]{0.68, 0.87, 0.83} 4.0 \footnotesize{$\pm$0.2} &\cellcolor[rgb]{0.68, 0.87, 0.83} 4.0 \footnotesize{$\pm$0.2} &\cellcolor[rgb]{0.74, 0.90, 0.86} 3.9 \footnotesize{$\pm$0.2} &\cellcolor[rgb]{0.61, 0.85, 0.79} 4.1 \footnotesize{$\pm$0.2} \\
    \bottomrule
    \end{tabular}
    }
    \caption{\textbf{Study 1 Quantitative Results.} Top: Absolute Objective A and B scores. Bottom: Likert scale responses to survey questions (1 = strongly disagree, 5 = strongly agree). Each cell: average across participants in each interface condition $\pm$ standard error.}
    \label{tab:study1_main_table}
\end{table}

\begin{figure*}
    \centering
    \begin{minipage}[t]{.27\textwidth}
        \subfloat[Edit Efficiency Analysis]
        {\label{fig:study1_edit_distance} \includegraphics[width=\textwidth]{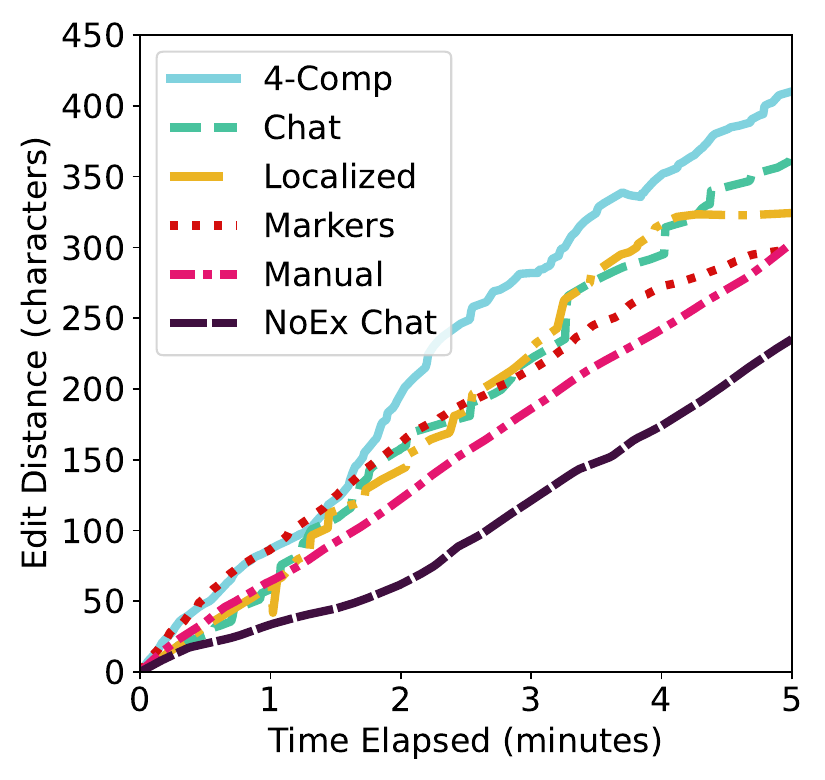}}
    \end{minipage}\hfill
    \begin{minipage}[t]{.33\textwidth}
        \subfloat[Recommendation Diversity]
        {\label{fig:study1_diversity} \includegraphics[width=\textwidth]{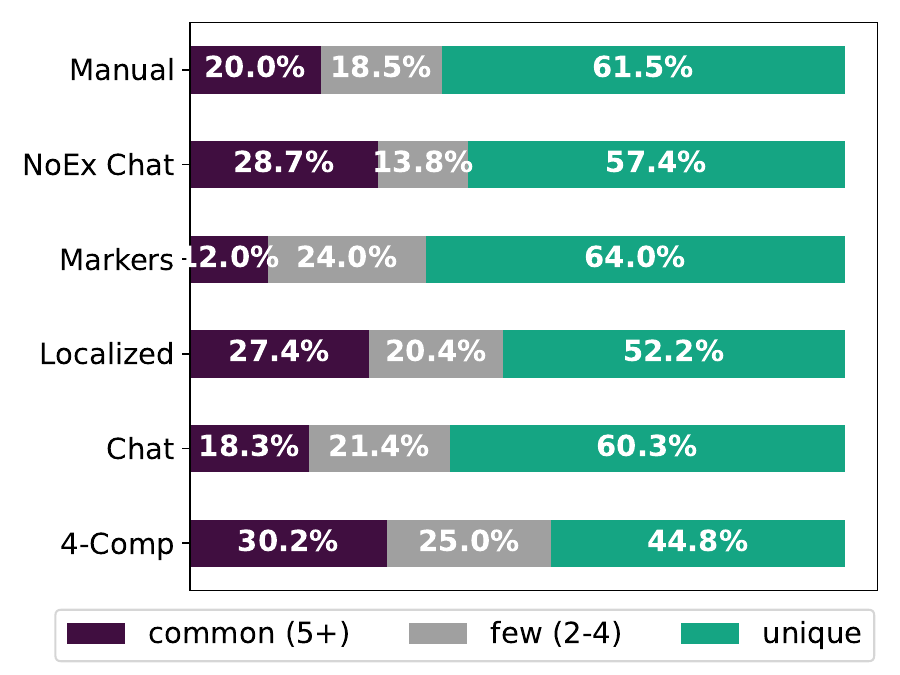}}
    \end{minipage}\hfill
    \begin{minipage}[t]{0.37\textwidth}
        \subfloat[Participant Rank Preferences]
        {\label{fig:study1_ranking} \includegraphics[width=\textwidth]{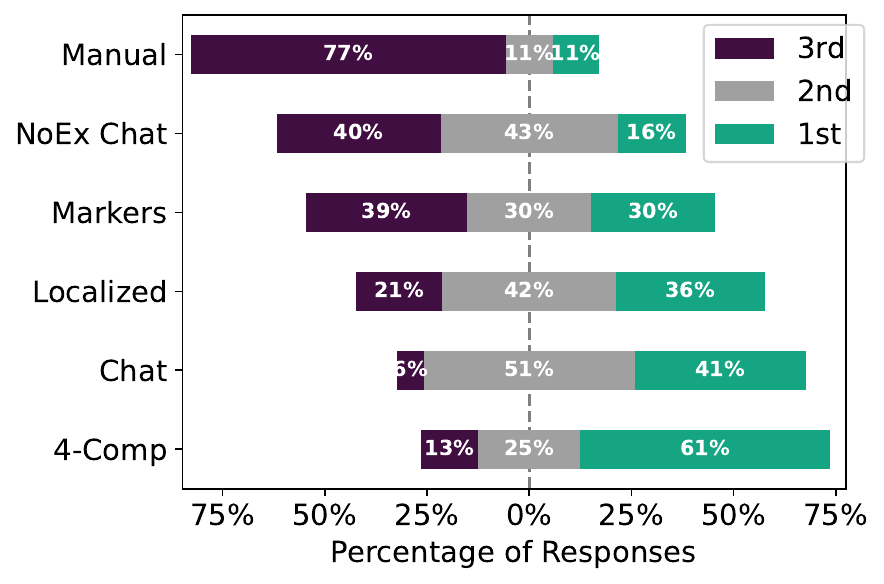}}
    \end{minipage}\hfill
    \caption{Study 1 Results: (\subref{fig:study1_edit_distance}) Average edit distance from the initial email template over time for each interface condition, (\subref{fig:study1_diversity}) distribution of recommendation diversity across interface conditions, (\subref{fig:study1_ranking}) User preferences for each interface condition.}
    \label{fig:study1_results}
\end{figure*}

We evaluate participants' completed sessions through multiple complementary lenses: (1) achievement of editing goal, (2) editing efficiency, (3) diversity of responses, (4) participant preference, (5) comparison of Chat and Comment conversations, and (6) factual accuracy.

To assess significance testing, we ran standard t-tests, applying the Bonferroni correction \cite{bonferroni1935calcolo} when multiple tests were performed on the same data. We also performed bootstrap re-sampling \cite{efron1982jackknife} to determine the minimum number of participants required to achieve statistical significance ($p < 0.05$). Details on statistical testing appear in Appendix~\ref{app:stats_significance}.

\subsubsection{Achievement of Editing Goal} \label{sec:study1_evaluation_goal}

We quantitatively measure participants' achievement of the two editing objectives based on the last version of the document at the end of the editing session. For the first objective -- ensuring the email has a formal tone and is professional -- we identify and count the number of typos, grammatical errors, and informal phrases in the email, and name this quantity \texttt{Objective A}. For the second objective -- customizing the email to the target destination and persona -- we identify and count the number of personalized suggestions related to the destination (e.g., suggesting the Eiffel Tower for a trip to Paris), or persona (e.g., suggesting Le Bernardin restaurant in New York for a food lover persona) in the email, and name this quantity \texttt{Objective B}.

We note that for Objective A, a lower quantity is preferred, and a higher quantity is preferred for Objective B. Annotation was completed in two stages: automatic annotation with an LLM, followed by manual verification (details on the annotation in Appendix~\ref{app:annotation_detail}).

Table~\ref{tab:study1_main_table} lists the average Objective scores under each interface condition. Focusing on Objective A, none of the interfaces lead to an ideal score of 0 (no detected issue) yet there are noteworthy differences. The two conditions that require manual editing (Manual, NoEx Chat) lead to the highest number of errors compared to the other four conditions (not significant, $p>0.05$), while interfaces with the Markers component (Markers, 4-Comp) achieve the best performance on Objective A, with roughly half as many errors as other conditions ($p<0.01$).

For Objective B, interfaces without a conversational component (Manual, Markers) lead to the least amount of customization, with 1-3 customizations on average, outperformed by other conditions with a conversational component ($p<0.01$), that all include 4-5 customizations.

Comparing the NoEx Chat and Chat conditions, executable edits slightly improve scores on both objectives, confirming that executable edits enable efficient integration of suggestions, and reduction of editing errors (not significant, $p>0.05$).

In summary, as shown in Table~\ref{tab:study1_main_table}, \textbf{conditions with the Markers component lead to significantly better writing quality (Objective A), and conditions with a conversational component (Chat, Comment) lead to significantly more customization (Objective B)}. The 4-Comp condition -- which has both Markers and conversational components -- leads to the best joint performance, supporting the additive benefits of the edit-suggesting components of InkSync.

\subsubsection{Editing Efficiency} \label{sec:study1_evaluation_efficiency}

We measure editing efficiency by recording the edit distance -- in the number of characters -- between the current version and the initial email template. This distance is recorded every few seconds during the editing session. This allows us to calculate the average edit distance over time for each interface condition, which represents how quickly participants diverge from the initial template.
Although editing efficiency does not explicitly measure success at the editing task, it indirectly measures the ease of editing to allow comparisons across each interface condition. Figure~\ref{fig:study1_edit_distance} plots the results of the analysis.

Participants in the NoEx Chat condition had the slowest editing speed overall ($p<0.05$), followed by the Manual condition. The slower editing in the NoEx Chat condition is probably caused by the need for participants to split their time and attention between the chat, and the manual integration of the chat's suggestions into the draft.

\textbf{All interfaces with components that suggest executable edits lead to faster editing than the manual condition} ($p<0.01$), demonstrating the usefulness of executable edits.
\textbf{The 4-Comp condition leads to the most efficient editing} ($p<0.05$), providing evidence of the components' complementarity in assisting the author's editing process.

\subsubsection{Diversity of Responses} \label{sec:study1_evaluation_diversity}

An important consideration of using LLMs to assist with writing tasks is whether they affect the creativity of the user. Creativity is often defined through the concept of \textit{divergent thinking} \cite{guilford1967nature}, in which a person or group generates multiple ideas or solutions to a problem; in this framing, creativity can be measured quantitatively by the number of distinct ideas generated \cite{torrance1966torrance}.

We use this idea to develop a quantitative method to measure the customization divergence in each interface condition. We manually cluster all customized recommendations extracted for \texttt{Objective B} and assign each recommendation to one of three classes: (1) \textit{common recommendations} are suggested in 5+ emails (e.g., the pyramid of Giza in Egypt), (2), \textit{uncommon recommendations} are suggested in 2-4 emails (e.g., a visit of the Luxor Temple in Egypt), and (3) \textit{unique recommendations} are suggested in only one email (e.g., a visit to the El Tahrir Square in Cairo).

In Figure~\ref{fig:study1_diversity}, we depict the recommendation class proportions for each study condition. We find that the Manual and Markers conditions -- ones with no conversational component -- yield the highest proportion of unique recommendations.

In other words, \textbf{we find evidence that the use of a conversational LLM in the Chat and Comment components leads to less original recommendations, a sign of reduced divergence in study participants, indicative of lowered creativity}. This finding is in accordance with recent work in argumentative essay writing \cite{padmakumar2023does}.

\subsubsection{Participant Preference} \label{sec:study1_evaluation_preference}

Participants completed a survey as the last step of the study. First, participants answered 9 Likert-scale questions about each of the three conditions they engaged with, with results presented in Table~\ref{tab:study1_main_table}. Second, participants ranked the three conditions; ranking results are compiled in Figure~\ref{fig:study1_ranking}.

The Manual and NoEx Chat conditions were ranked the lowest ($p<0.01$), and Likert answers indicate participants were least likely to agree that those systems are helpful in completing the task and provide easy-to-integrate suggestions. 

Participants preferred the 4-Comp condition and the Chat condition most (no significant difference between these two), with Likert responses indicating that system suggestions were the easiest to understand, adjust, and integrate.

Surprisingly, participants felt it was equally easy to specify the desired location of an edit in the Comment and Chat conditions (no statistical difference). One hypothesis for this lack of a difference is that the emails are relatively short documents and that the benefits of precisely localized conversations would only materialize in longer documents.

\subsubsection{Chat vs. Comment Conversations} \label{sec:study1_chat_v_comment}

The Chat and Comment components each provide a conversational interface to the participant. In Appendix~\ref{app:chat_vs_comment}, we describe an analysis we performed on all the Chat and Comment conversations of Study 1 participants. Figure~\ref{fig:example_chat_comment} provides illustrative Chat and Comment conversations from participants in the study, and Table~\ref{tab:chat_vs_comments} summarizes key qualitative differences.

At a high level, \textbf{we observe important differences in participant's Chat and Comment conversations.} Chat conversations are longer (8.6 vs. 3.0 messages), and have author queries with more diverse intents; they request edits of local and global scope while not usually specifying localization explicitly. This causes the system to have to implicitly resolve ambiguity and respond with a wider variation in the number of suggested edits. By contrast, Comment conversations are shorter, with participants predominantly requesting edits of local scope. The Comment component frequently suggests a single edit, representing a much more focused and efficient localized conversation.

\subsubsection{Factual Accuracy} \label{sec:study1_evaluation_accuracy}

The final evaluation centers on the factual accuracy of the information added by participants to the emails. We reviewed each email and identified any recommendation that was not accurate for the target destination or persona, identifying three categories of error: (1) overselling (e.g., claiming that Paris is an affordable city), (2) destination-related errors (e.g., recommending to visit a zoo that permanently closed in 2017), (3) persona-related errors (e.g., recommending a bike tour for a person with limited mobility). Besides the error category, we manually attributed a severity to each identified error of Minor or Major, following the MQM annotation format \cite{freitag2021experts}. Appendix~\ref{app:study1_accuracy} details the annotation procedure, with Table~\ref{tab:study1_inacurrate_examples} listing examples of each error category and severity, and Table~\ref{tab:study1_accuracy_breakdown} breaking down error counts by study condition.

A key limitation of this analysis is our reliance on automation for error detection, possibly resulting in undetected factual errors and underestimating the actual number of factual errors in the documents.  Thus the error percentages reported here should be considered a lower bound.

Factual errors occur in all conditions (including the manual interface condition). \textbf{However, major errors are slightly more likely to occur in a condition with a conversational interface with 14.8\% of documents containing a major error}, compared to 6\% for non-conversational interfaces (Manual, Markers). In summary, we find evidence that recommendations made by the system in conditions with conversational components lead to a higher likelihood of the participant integrating a major factual error in the email.

In the following study, we assess the efficacy of the Warn-Verify-Audit framework to reduce factual errors.

\section{Usability Study 2: Warn, Verify and Audit} \label{sec:study2}

We conducted a 20-minute within-participants usability study with 35 participants, based on the same task as Study 1, but modified to increase the chance of system-generated inaccuracies, enabling us to evaluate the Warn, Verify, and Audit components of InkSync. The objectives were to:
\begin{enumerate}
    \item Determine whether the Warn, Verify and Audit components can help authors detect factual errors and avoid integrating them into their documents,
    \item Assess whether the search engine queries produced by the Verify component are adequate and sufficient for users to verify the suggested edit's accuracy,
    \item Understand the differences between verifying information during the editing phase versus the auditing phase.
\end{enumerate}

\subsection{Participants} \label{sec:study2_participants}

We recruited participants from our initial survey respondents and added referred participants to the study by initial participants. Sixteen days elapsed between the end of Study 1 and Study 2's start and all participants were provided with the necessary introductory material to complete the study.

\subsection{Modifying the System to Introduce Errors} \label{sec:study2_modifications}

Since the focus of Study 2 is to evaluate the Warn, Verify, and Audit components, we removed the Markers, Comments, and Brainstorm components. The Chat component was the only way for participants to receive system suggestions.

Since Study 2 assesses the helpfulness of system affordances in identifying errors, it is important to ensure that some errors occur in all editing sessions.
To do so, we created a modified version of the Chat prompt, which we call the perturbed prompt, that is used to encourage the model to suggest inaccurate information. This is achieved by explicitly instructing the model to introduce subtle factual errors, and giving it examples of such errors (selected from Major errors identified in Study 1).

During editing tasks in Study 2, the system alternates between the standard and perturbed prompts when responding to user Chat queries, aiming for the system to suggest an equal number of factually accurate and inaccurate edits on average. We manually verified that the perturbed prompt was effective in causing the model to suggest inaccurate edits and show examples of such edits in Appendix~\ref{app:study2_inaccuracies}.

\subsection{Study Procedure} \label{sec:study2_procedure}

Study 2 follows the same format as Study 1, with three 6-minute tasks framed around email customization, followed by a 2-minute completion survey. Study 2 consists of three tasks:

\begin{enumerate}
    \item Task 1: The participant customizes an initial email template without interface support for verification. 
    \item Task 2: The participant customizes an initial email template with the Warn and Verify components enabled.
    \item Task 3: The participant audits an email edited by a participant in Task 1 or 2, in the Audit interface. 
\end{enumerate}

For the editing tasks of Study 2, we reused the same fictional recipient names and personas from Study 1 but selected three new target destinations -- Madagascar, Jordan, and Malaysia -- focusing on destinations likely to be less familiar to US-based participants, increasing participants' reliance on the system for recommendation and verification.
In order to ensure that there is erroneous content to audit in the third task, we selected an email edited by a participant in the first two steps of the study with at least five integrated system edits, including both accurate and inaccurate edits.

During the five-minute auditing task, the participant was instructed to verify all system-generated content in the document, marking each with a ``Verified'', ``Incorrect'', or ``Not Sure'' label. If the participant cannot complete all verifications in the allotted time, unmarked edits are assigned a ``Not Enough Time'' label for analysis purposes.

\subsection{Evaluation Methodology and Results} \label{sec:study2_evaluation}

\begin{figure*}
    \centering
    \begin{minipage}[c]{.35\textwidth}
        \subfloat[Statistics of the Verification Process]
        {\label{fig:study2_verif_stats} \includegraphics[width=\textwidth]{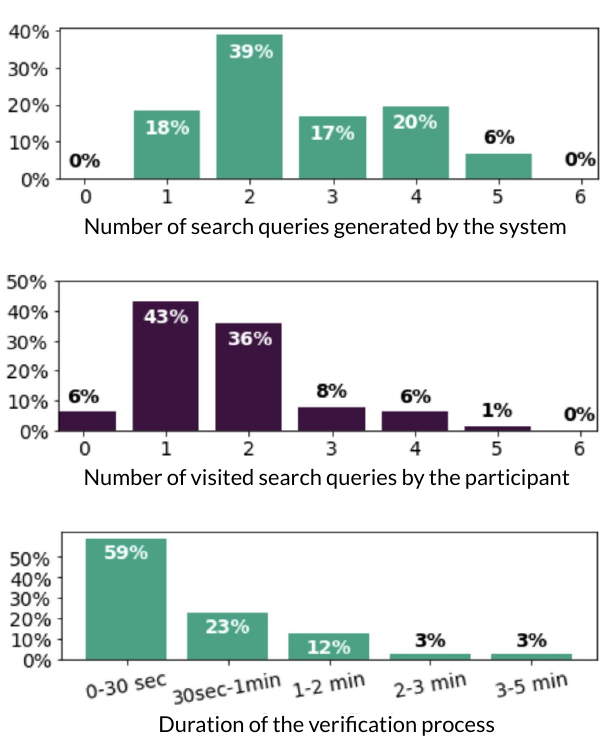}}
    \end{minipage}\hfill
    \begin{minipage}[c]{0.35\textwidth}
        \subfloat[Study 2 Participant Feedback]
        {\label{fig:study2_feedback} \includegraphics[width=\textwidth]{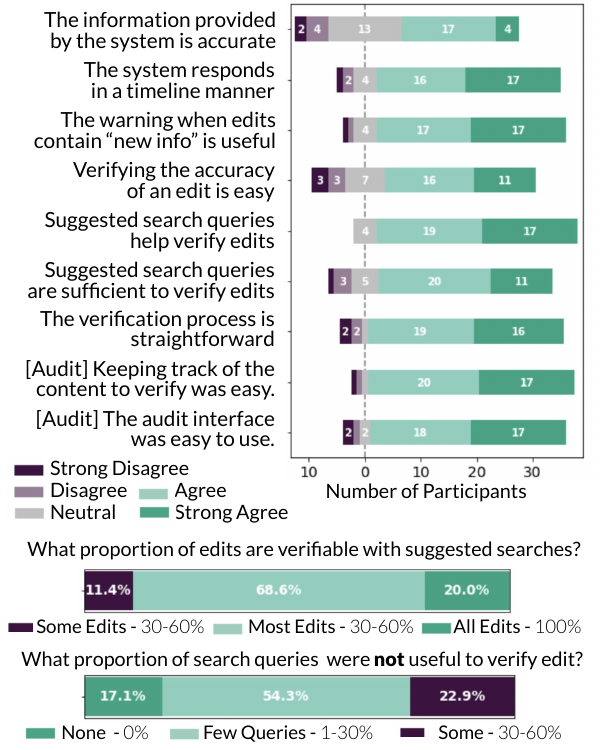}}
    \end{minipage}\hfill
    \begin{minipage}[c]{.26\textwidth}
        \subfloat[Preventing and Detecting Errors]
        {\label{fig:study2_diagram} \includegraphics[width=\textwidth]{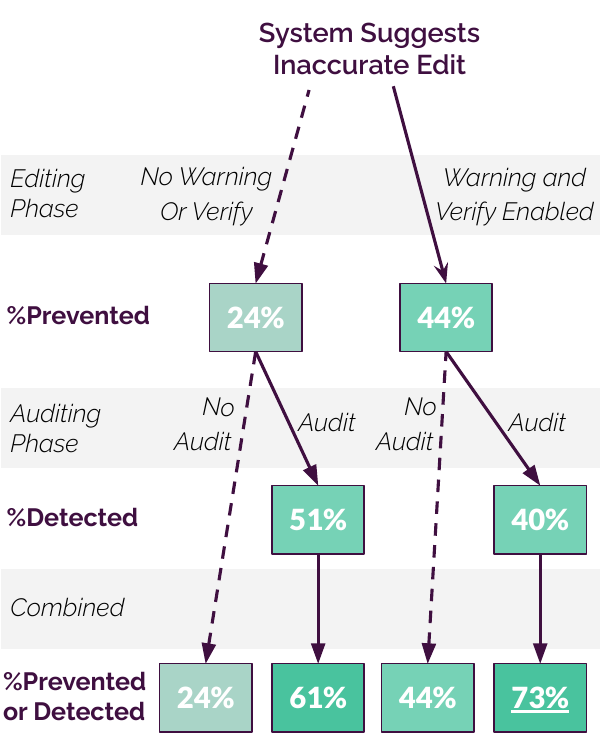}}
    \end{minipage}\hfill
    \caption{Study 2 Results: (\subref{fig:study2_verif_stats}) Statistics of the 500+ verifications conducted by participants, (\subref{fig:study2_feedback}) User responses from Study 2's final survey, (\subref{fig:study2_diagram}) Summary of proportion of errors prevented and detected in varying conditions of Study 2}
    \label{fig:study2_results}
\end{figure*}

We evaluate Warn, Verify, and Audit framework through statistics of the verification process, participant feedback, and participant success at detecting and preventing inaccuracies.

\subsubsection{Statistics of the Verification Process} \label{sec:study2_evaluation_verification}

We track participant interactions with the Verify component and record relevant statistics, summarized in Figure~\ref{fig:study2_verif_stats}. We report: (1) the \textit{number of generated search queries} by the system when verification is initiated, (2) the \textit{number of visited search queries} by participants from the suggested list, and (3) the \textit{duration of the verification process}: the time elapsed between verification initiation and the participant assigning an accuracy label to the edit.

In the 500+ verifications initiated by participants, the system generated between one and six search queries, and 2.6 on average (distribution in Figure~\ref{fig:study2_verif_stats} (top)). Participants visited 1+ recommended searches in 94\% of cases and on average 1.67 (Figure~\ref{fig:study2_verif_stats} (middle)), and on average visited 70\% of recommended searches, suggesting that these offerings were often found helpful. On average, participants spent 44 seconds performing a verification, and 80\% of the verifications were completed in under a minute, providing empirical evidence that the system enables participants to conduct information verification efficiently (distribution in Figure~\ref{fig:study2_verif_stats} (bottom)).

\subsubsection{Participant Feedback} \label{sec:study2_evaluation_feedback}

The completion survey consisted of nine Likert-scale statements and two multiple-choice questions about various aspects of the Warn, Verify, and Audit components. The statements, questions, and aggregated responses are shown in Figure~\ref{fig:study2_feedback}.

Regarding the accuracy of the system overall, participants were unlikely to agree that the information provided by the system was accurate (average 5-point Likert of 3.2). This suggests that participants were aware of the system-introduced factual errors during the study.

Regarding the verification process, most participants found the ``new information'' warning to be useful (4.0) and thought that the verification process was straightforward (4.3). When asked about the suggested search queries, participants were very likely to agree that the search queries are helpful (4.3), suggesting that this is a viable method for helping verify information while editing. However, the average is lower for the sufficiency of the search queries (3.8). This gap suggests that fully completing a verification might require adapting or expanding the system-suggested searches (not significant $p>0.05$).

Most participants were positive about the auditing interface, agreeing that it helped keep track of content that required verification (4.2) and that the auditing interface was easy to use (4.0).

The final two questions asked participants to quantify their impressions of the verification process. When asked what proportion of the \textit{edits} were verifiable with the suggested search queries, 85\% stated that most or all edits were verifiable.
When asked what proportion of the \textit{queries} were \textit{not} useful for verifying edits, most participants stated <30\% of search queries were not useful in the verification process. In summary, most of the suggested search queries were useful in the verification process and the system enabled participants to verify most edits that they chose to review.

\subsubsection{Inaccuracy Prevention and Detection} \label{sec:study2_evaluation_inaccuracy}

\begin{table}[]
    \centering
    \begin{tabular}{p{5.3cm}cc}
    
    \cmidrule{2-3}
    & \textbf{Task 1:} & \textbf{Task 2:} \\
    \textbf{Session Statistic} & \textbf{No Warn/Verif} & 
    \textbf{Warn/Verif Enabled} \\
    \midrule
    Avg. Messages in Chat Conversation & 13.7 & 16.0 \\
    Avg. Accurate Edits Suggested & 2.5 & 3.1 \\
    Avg. Inaccurate Edits Suggested & 3.0 & 3.1 \\
    Avg. Initiated Verifications & 0.0 & 2.1 \\
    Avg. Inaccurate Suggestions Accepted & 2.3 & \textbf{1.7} \\
    \% Inaccurate Edits Prevented & 23.4 & \textbf{43.5} \\
    \bottomrule
    \end{tabular}

    \caption{Effect of the Warn and Verify component in Study 2 for participants completing an editing task.}
    \label{tab:study2_step1-2}
\end{table}

Table~\ref{tab:study2_step1-2} summarizes key statistics of the two editing sessions -- Tasks 1 and 2. Recall that Task 2 provided Warning and Verification support, while Task 1 did not. On average, participants had longer exchanges with the system in the Chat in Task 2 than in Task 1 (16.0 vs. 13.7 messages exchanged), leading to a larger number of edits suggested to participants in Task 2 than in Task 1 (6.2 vs. 5.5 suggested edits). In both conditions, the system recommended a similar proportion of accurate and inaccurate suggestions, as intended in the study's design.
We calculate a quantity called \texttt{\%Inaccurate Edits Prevented}, which is the percentage of inaccurate suggested edits that participants dismissed and did not integrate into the final email. For instance, if the system suggests 5 inaccurate edits, and the participant dismisses 3 of them and 2 remain in the current draft, \texttt{\%Inaccurate Edits Prevented} is 60\%.

Participants in Task 1, which did not include support for Warning and Verification, prevented only 23.4\% of inaccurate suggestions (accepted 2.3 of the 3.0 inaccurate edit suggestions). Participants in Task 2, which did have support for Warning and Verification, prevented 43.5\% of edits from being accepted into the draft (accepted 1.7 of the 3.1 inaccurate edit suggestions). \textbf{In other words, enabling the Warn and Verify features almost doubles the proportion of inaccurate edits that are successfully dismissed by the participant ($p<0.05$).}

For the auditing task -- Task 3 -- we calculate a quantity called \texttt{\%Inaccurate Edits Detected}, which is the percentage of inaccurate edits in the audited document that are marked as ``Incorrect'' by the participant. In other words, if the audited document contains a total of four inaccurate edits and the participant marks three of them as ``Incorrect'', \texttt{\%Inaccurate Edits Detected} is 75\%.

In Study 2, documents were either edited with or without verifications enabled (Task 1 vs. Task 2) and the document was either audited or not (Task 3), leading to four possible conditions. Figure~\ref{fig:study2_diagram} provides statistics for documents in these four conditions, and the bottom row indicates the percentage of inaccurate edits that were either prevented or detected for each condition. Documents that were edited with verification enabled, and audited in a secondary phase led to the highest rate of error avoidance at 73\%. \textbf{This finding confirms the additive benefits of providing verification during editing and a separate auditing phase}, as it leads to the largest fraction of prevented and detected errors ($p<0.01$), more than three times as many as in the condition with neither in place.

\section{Discussion} \label{sec:discussion}

\subsection{Implications}

Despite their recent introduction, and although they are known to have many drawbacks, chatbot-style LLMs are both extremely powerful and widely used for knowledge work today. The research presented here shows that taking a human-centered stance that emphasizes human agency and control can lead to better outcomes. The interface design and workflow of InkSync allow an author to see precisely which LLM suggestions have been made and how they integrate into the text.

The results of the two usability studies shed light on the details of this form of human-AI interaction. We found that specialized query tools (the Markers), such as grammar checkers and tone adjusters, can lead to more accurate edits than open-ended queries alone. On the flip side, we found that open-ended queries (the Chat and Comment interactions), lead to more customization, as in a higher number of suggested places to go and things to do for a given persona and travel destination. At the same time, there was more redundancy in the suggestions made in the open-ended chat conversations than in the Manual and Markers conditions. In other words, people writing manually were more creative in the content they produced, but they produced less new content. 

We found that open-ended conversation tools without verification controls lead to more factual errors than the same tools with scaffolded support for verification and fact-checking. Our research shows a promising solution for how to provide this scaffolding: the Verify and Audit facilities help make the author aware of potentially problematic content generated by the LLM, and reduce the friction of repairing introduced factuality errors.

We also found that participants prefer using open-ended, controlled conversational LLMs over the other options, underscoring the point that because people like these tools, researchers need to determine how to provide scaffolding to increase their accuracy.

Altogether, our results provide guidance for the development of future LLM-based editing tools. The InkSync software will be provided as open-source software, expanding the potential impact of this work by enabling other researchers to build upon it.

\subsection{Comparison to Existing Commercial Tools}

The interface design of the Markers component bears resemblance to modern text checkers such as Grammarly\footnote{https://www.grammarly.com/}, which proactively detect errors and suggest corrections for a limited set of specialized tasks. Results from Usability Study 1 confirm the effectiveness of such tools in helping authors improve the quality and tone of their writing but are limited by the inability of the author to specify suggestions other than those that are pre-specified in the tools. Our design of Markers, paired with the use of LLMs as the backend correction tool, provides control to the author to define the scope of suggestions they want, based on their editing objectives. For example, when writing a social media post, the author might need markers that can work to help improve Engagingness or suggest Emojis to insert, while a teacher working on simplifying text for a classroom setting might need markers to automatically provide definitions to complex terms, split sentences, or suggest lexical substitutions.

Popular co-writing tools such as Google Docs\footnote{https://docs.google.com/} and Overleaf\footnote{https://www.overleaf.com/} integrate comment features that are similar in design to the Comment component of InkSync. As the commercial tools inevitably integrate LLM-based suggestions, our findings from Usability Study 1 on the use of Comments for focused localized editing indicate that comments are a good place for LLM-based interaction. In a multi-author setting, however, the LLM would need to follow co-writing social norms and understand when it is appropriate to suggest edits to an author, and when a conversation only involves the human authors.

The Auditing interface of InkSync shares features with revision history tools such as Track Changes in Microsoft Word, or the version histories of Overleaf and Google Docs. A key distinction is that the auditing in InkSync centers around the latest version of the document focuses specifically on LLM-generated content, and provides verification functionality.  
By contrast, revision history tools focus on the evolution of the document over time and are designed to help authors understand the changes made by other (human) authors. As system-generated content makes its way into commercial writing tools, revision history tools will need to evolve to support both revision history and auditing of system-generated content.

\subsection{Limitations} \label{sec:limitations}

Studies 1 and 2 showed that providing scaffolding in an LLM-based editing interface has many benefits over a standard conversational interface. However, there are some potential limitations to the breadth of the applicability of these results. The participant population consisted mainly of US-based English-speaking knowledge workers; the results found here might not transfer to other populations. The studies were of short duration, whereas longitudinal studies that show changes in use over time can reveal different usages over time. The evaluations assessed only short (100-400 word) documents, both to reduce the duration of study sessions and to avoid LLM latency issues. Assessing on longer documents might reveal new insights. 

Although InkSync works with any LLM (i.e., components in InkSync that interface do not require a particular LLM to function), we only experimented with OpenAI's GPT-4 model to conduct the usability studies. We selected GPT-4 as it was a top-performing language model at the time of our studies, indicated by high performance on common benchmarks such as MMLU \cite{hendrycks2020measuring}. High-quality model outputs are essential to evaluating the usability of the system, as prior work has often found that low-quality system suggestions lead to degraded participant activity \cite{clark2018creative}. Yet relying on GPT-4 -- a closed source model accessible only through API -- is limiting: OpenAI continuously updates the model served as GPT-4, hampers our findings' reproducibility and the lack of details on the training of GPT-4 prevents the community from building knowledge on the model's limitations. Future work can integrate other closed and open-source LLMs into InkSync, and measure the underlying model's effect on the user's interaction.

\subsection{Future Directions}

This work can be expanded in several directions:
\paragraph{Increasing Response Diversity with the Temperature Parameter.} In the study of Section~\ref{sec:study1}, we find that LLM-based suggestions tend to limit participants' creativity in their email customizations. By default, language models aim to optimize generated text's likelihood, which can lead to text that is trite and generic \cite{arnold2016suggesting}. However, language models have controllable parameters such as temperature, which can be used to encourage the model to generate more diverse content \cite{roemmele2018automated}. However, this setting is difficult to control, and exploring its use for editing suggestions is an area for future work.

\paragraph{Personalized Interaction.} In the current implementation of InkSync, the system ignores past user interactions (e.g., the author accepting or dismissing an edit). The system only relies on the document's current state and active author queries to generate edits, which is a `memoryless' approach. Since prior work has shown LLMs have the ability to mimic an author's style \cite{reif2022recipe}, future work could explore personalization by integrating the author's interaction history, including in previous documents.

\paragraph{Expanded Verification Framework.} In the current implementation of InkSync, the information verification framework assesses the accuracy of system-generated content. Yet Study 1 findings indicate that inaccurate content also appears in the fully manual text editing interfaces, indicating that human-written text can also contain inaccurate information. Future work could expand the verification framework to probe the accuracy of \textit{all} content in the document irrespective of its provenance. Through the use of frameworks like ToolFormer \cite{schick2023toolformer}, LLMs can make use of tools such as a search engine or knowledge base to perform a preliminary verification for authors. Automated verification would likely require proof-of-work with traceable citations \cite{gao2023enabling}, currently still challenging for LLM-based systems such as Bing Chat \cite{liu2023evaluating}.

\paragraph{LLMs in collaborative writing tools.} InkSync currently supports a single author with no concurrent editing features. Yet collaborative editors (e.g., Google Doc, Overleaf) are now widely used. \cite{brodahl2011collaborative}. Adapting the LLM-based features of InkSync to collaborative writing interfaces is a challenging task, both to track the different authors' intent and because collaborative environments involve complex social norms \cite{reagle2010nice, gero2023social} that the LLM might have to adopt.

\section{Conclusion} \label{sec:conclusion}

In this work, we presented InkSync, a text editing interface with LLM-based components that can suggest transparent and executable edits to human authors as they write. InkSync also implements a three-stage information verification framework -- Warn, Verify, Audit -- to warn authors about possible factual errors, verify edits, and track content origin in the final document, simplifying the process of reviewing the accuracy of auto-generated passages. A first usability study revealed that the complete interface which includes four edit-suggesting components improves the efficiency and effectiveness of participants at editing emails over baselines such as manual editing, or single-component interfaces. In a second study, we confirm the effectiveness of the verification framework, finding that it helps participants detect and avoid three times as many factual inaccuracies compared to an interface without the framework implemented. We hope that InkSync and the empirical insights from our two usability studies can inspire future research and products into writing tools that offer advanced LLM interaction while providing the tools for authors to ensure content accuracy and mitigate the propagation of misinformation.


\bibliographystyle{ACM-Reference-Format}
\bibliography{biblio}

\appendix

\section{Appendix}

\subsection{LLM Prompt for Chat Component} \label{app:llm_chat_prompt}

\begin{figure*}
    \centering
    \begin{minipage}[t]{0.49\textwidth}
        \subfloat[Prompt guideline used to operate the Chat component.]
        {\label{fig:chat_prompt_guidelines} \includegraphics[width=\textwidth]{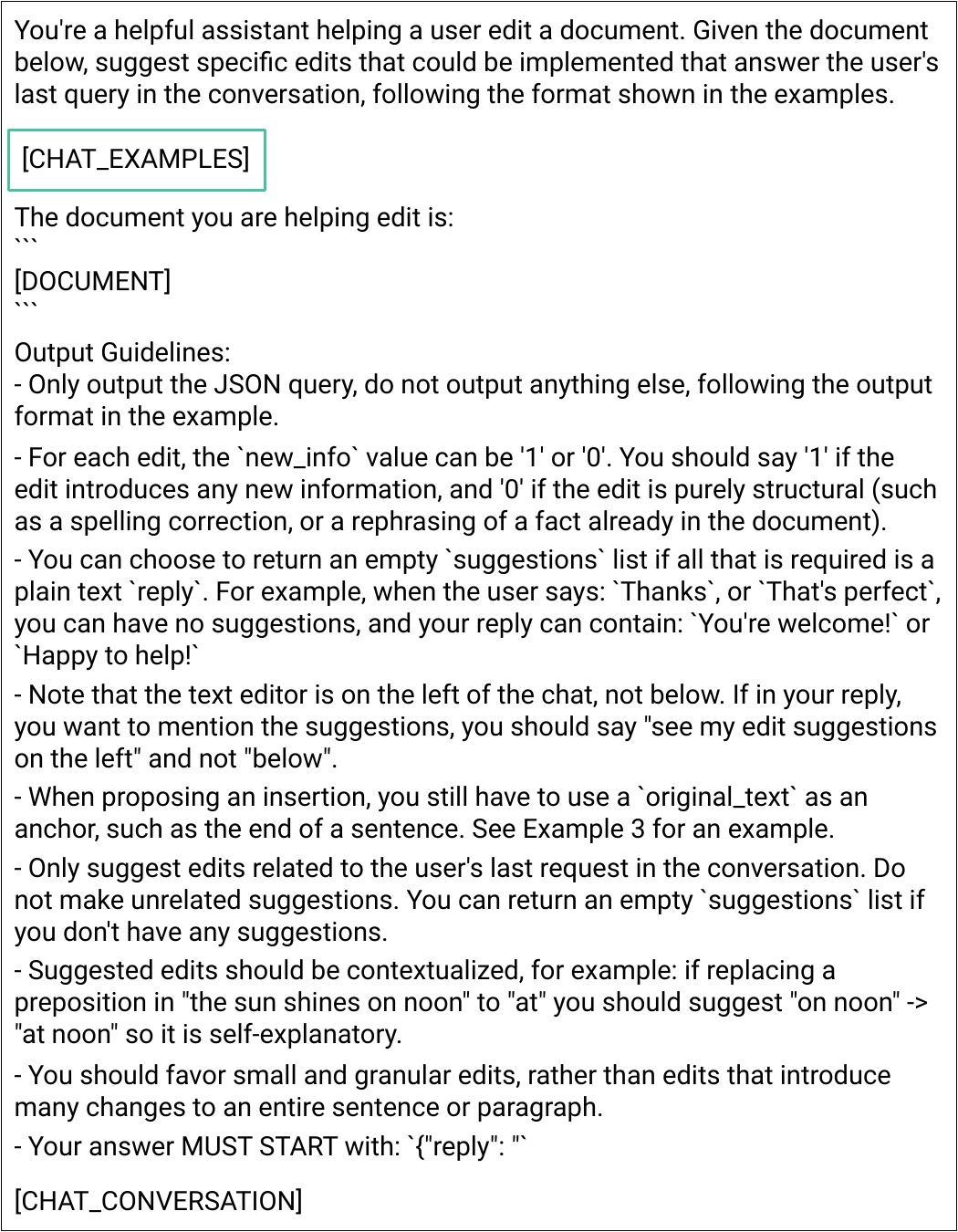}}
    \end{minipage}\hfill
    \begin{minipage}[t]{.49\textwidth}
        \subfloat[Three examples of expected output for the Chat component.]
        {\label{fig:chat_prompt_examples} \includegraphics[width=\textwidth]{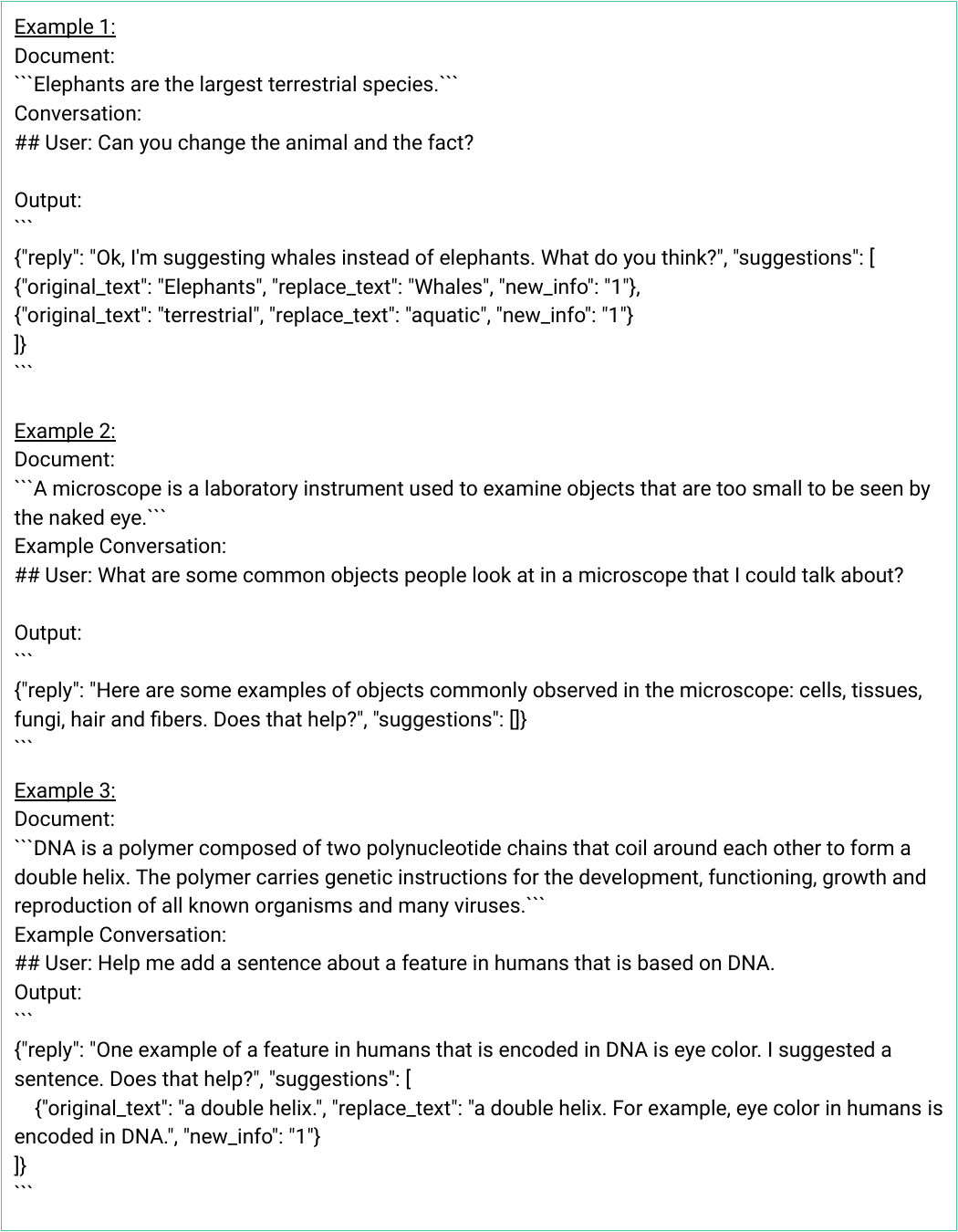}}
    \end{minipage}\hfill
    \caption{The prompt used to operate the Chat component. The complete prompt is formed by inserting the three examples in (\subref{fig:chat_prompt_examples}) into the (\subref{fig:chat_prompt_guidelines}) guidelines at the \texttt{[CHAT\_EXAMPLES]} location, the latest version of the document is inserted into the \texttt{[DOCUMENT]} location, and the conversation containing author queries in \texttt{[CHAT\_CONVERSATION]}.}
    \label{fig:chat_prompt_full}
\end{figure*}

Figure~\ref{fig:chat_prompt_full} details the prompt used to operate the Chat component (Section~\ref{sec:chat}). The complete prompt is composed by inserting the three examples of expected output into the \texttt{[CHAT\_EXAMPLES]} slot of the prompt guideline, the latest version of the document in the \texttt{[DOCUMENT]} slot, and the conversation containing author queries and system replies in \texttt{[CHAT\_CONVERSATION]} slot. The listed output guidelines provide the Design Choices to the LLM as well as additional formatting information which was added iteratively to minimize system errors (e.g., your answer must start with [...]).

\subsection{Character-Level Provenance Tracing Algorithm} \label{app:tracing_algorithm}

In order to provide precise content provenance in the Audit interface, we implement a character-level tracing algorithm for any system-generated content that is accepted by the author during the editing process.

First, note that the acceptance of an edit during the editing process does not guarantee that the edit will be present in the final version of the document. In a subsequent action, the user might remove the edit partially or entirely, or the edit's content might not remain contiguous (e.g. if the user adds some text in the middle of the edited text). Accurate content provenance tracing therefore requires implementing a character-level algorithm to track system-generated content through the editing process.

In our implementation, manually typed characters are marked as user-originated, and any time a system suggestion is accepted into the draft, all characters inserted into the document based on this suggestion are marked as system-originated, and linked to the specific edit. The state of the document is saved periodically as the user edits the document. Each time the document is saved, we use the character-level Levenshtein string alignment algorithm \cite{Levenshtein1966BinaryCC} to track the movement of each system-generated character and record its new position in the latest document version.

Once an auditor requests to view the content provenance of a document, character-level tracing information is aggregated into contiguous spans and mapped back to originally accepted edits.

\subsection{Study Task: Email Customization Options} \label{app:study_customization_options}

Participants in both usability studies were tasked with editing email templates for the fictional InkSync travel agency. To make the task more specific, participants were provided with a target destination, and a customer persona, randomly selected (without replacement when participants completed multiple editing tasks) from pre-selected options.

For Study 1, the five target destinations were: Paris, New York, Argentina, Singapore, and Egypt. The four customer personas were: a business traveler who loves good food, a student on a budget, a parent with two 6-10-year-old children, and a retired person with limited mobility.

Since participants completed three editing tasks in Study 1, we ensured that each participant saw a new target destination and persona for each task.

For Study 2, there were two editing tasks (Task 1 and 2), and we selected three new target destinations: Madagascar, Malaysia, and Jordan. Note that for this second study, we purposefully selected destinations that are less well-known to participants living in Western countries, to ensure that participants would rely on the system recommendations to customize emails, rather than their own knowledge of the target destination. Regarding the customer personas in Study 2, we selected the same four personas as in Study 1.

Below, we provide a concrete example of the shortest initial email template that participants edited during editing tasks. We note that this example is the shortest initial prompt, the other templates are not included in the Appendix for brevity considerations but are available in the open-source repository. 

\begin{myjsonblock}{Initial Email Template - Study 1 (1 of 3)}
Subject: Plan your next vacation.\\

Hey [Customer's Name],\\

How have you been? Are you tired and wanna go on a vacation? I have special invitation to [City Destination], here are the details.\\

Experience vibrant cultur and explore the stunning landscapes the place has to offer.\\

Dont miss out on this fantastc opportunity! Reach out to us now to book your escape. Your dream vacation awaits!\\

Cheers,\\
InkSync Agency
\end{myjsonblock}

\subsection{Study 1: Quality Control} \label{sec:study1_quality_control}

Since participants completed  Study 1 unmoderated, we implemented additional controls in the interface to help ensure the validity of the findings.

First, we disabled pasting of more than 50 characters into the text editor, to discourage participants from copying large amounts of text from an external source (e.g., Google search, an external LLM interface, etc.).

Second, participants were instructed to complete the study on a laptop or desktop, and the interface displayed an error message if the window size was below the size of a 13'' screen at standard resolution. Third, the interface was simplified: we removed the document list panel and used the space gained to provide study-related information such as the target destination, target persona, and a link to access the introductory material if needed.

Third, we added code to our interface disabling browser extensions such as Grammarly\footnote{https://stackoverflow.com/questions/37444906/how-to-stop-extensions-add-ons-like-grammarly-on-contenteditable-editors}, to ensure participants solely relied on the tools provided in the study condition, rather than support tools available by default in their browser.

Fourth, participants were instructed to complete the editing task in 5 minutes and were warned within the interface once the allotted time had elapsed, but participants were allowed to continue editing to reach a desired milestone prior to moving on to the next task. We verified that less than 10\% of participants spent more than 6 minutes completing a task, ensuring fair comparison across interface conditions.

\subsection{Study 1: Example Edited Email} \label{app:study1_example}

\begin{figure}
    \centering
    \includegraphics[width=0.7\textwidth]{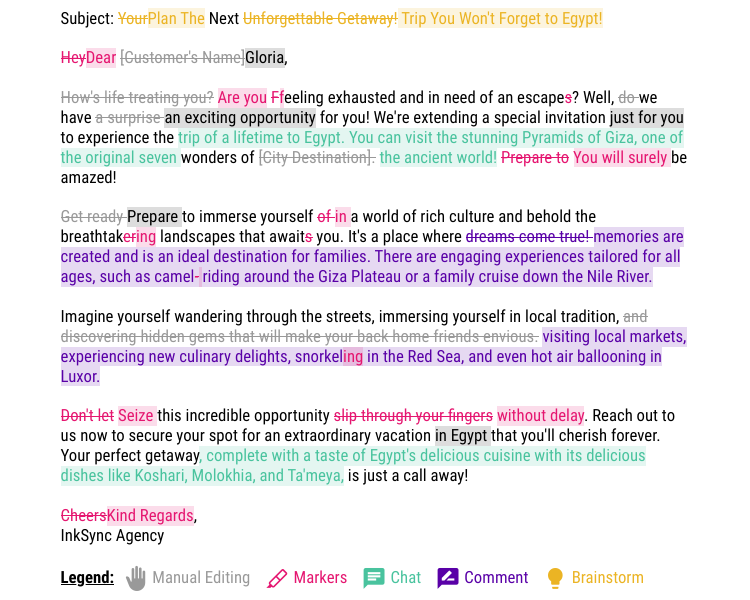}
    \caption{Example email edited by a participant of Study 1 in the 4-Comp condition (which includes the four edit-suggesting components: Markers, Chat, Comments, and Brainstorm). Background color indicates inserted text during editing, whereas \sout{strike-through} indicates deleted text. The color indicates the component used to accomplish the editing: Manual editing (grey), Markers (pink), Chat (green), Comment (purple), and Brainstorm (yellow).}
    \label{fig:study1_email_example}
\end{figure}

Figure~\ref{fig:study1_email_example} provides a concrete example of the editing outcome of a participant from Study 1. For this task, the participant was in the 4-Comp condition and had access to the four edit-suggesting components of InkSync. The participant manually edited portions of the email (shown in grey), and used each of the four components for other edits, which are shown through color-coding.

The Markers and Brainstorm components were mostly used by the participant to improve language and remove grammatical errors, while the Chat and Comment features were used to expand on paragraphs, customizing the email with information specific about Egypt (the target destination), and the persona (a parent with two children aged 6-10).

\subsection{Study 1: Annotation Detail} \label{app:annotation_detail}

In order to understand participant interaction, we performed several annotations on data collected in Study 1. Due to the large number of emails edited by participants (around two hundred), we implemented a two-stage annotation procedure. We first wrote detailed few-shot prompts for the annotation and prompted GPT-4 to obtain initial labels for the analysis. We then manually reviewed all labels, and updated labels when required to ensure the high quality of the annotations.

Although the addition of GPT-4 to the assessment process introduces some level of error, we expect the error to be consistent across all interface conditions, and therefore not affect the relative differences in the comparison across conditions. We will release all prompts used for annotation with the release of the data, to facilitate reproducibility.

\subsection{Study 1: Chat vs. Comment Conversations} \label{app:chat_vs_comment}
\begin{figure}
    \centering
    \includegraphics[width=0.6\textwidth]{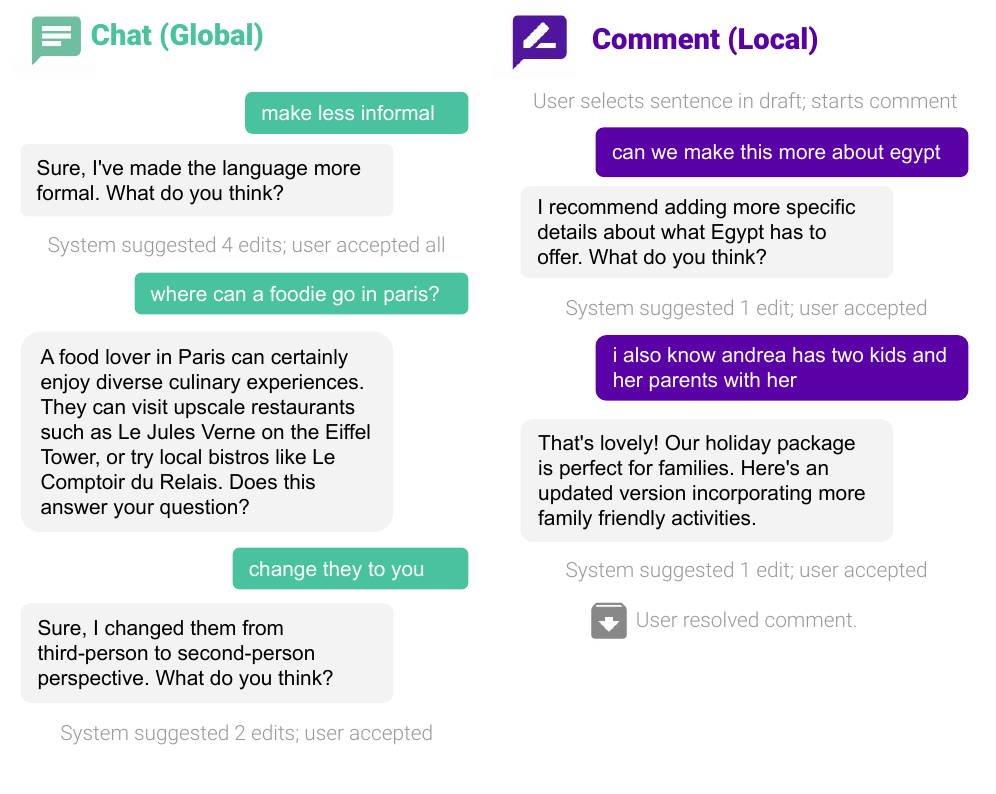}
    \caption{Illustrative examples of Chat and Comment conversations by participants in Study 1. Chat conversations tend to be longer, have more ambiguity, and require more diverse system responses. Participant messages are in green (chat) and purple (comment) boxes and system responses are in grey boxes.}
    \label{fig:example_chat_comment}
\end{figure}

\begin{table}[]
    \centering
    \resizebox{0.5\textwidth}{!}{%
    \begin{tabular}{lrr}
     & \symbolimg{figures/icons/chat.png} \textbf{Chat} & \symbolimg{figures/icons/comment.png} \textbf{Comment} \\
    \midrule
    Sample Size & 60 & 97 \\
    Avg. Conversation Length & 8.55 & 2.97 \\
    \midrule
    \multicolumn{3}{c}{
    \cellcolor[rgb]{0.95, 0.95, 0.95}
    \textbf{Distribution of User Message Classification}}\\
    \midrule
    \%Edit Request & 36.32 & 59.03 \\
    \%Information Seeking & 41.26 & 27.08 \\
    \%Chit Chat & 22.42 & 13.89 \\
    \midrule
    \multicolumn{3}{c}{
    \cellcolor[rgb]{0.95, 0.95, 0.95}
    \textbf{Scope of Edit (for Edit Request Only)}}\\
    \midrule
    \%Local Edit Request & 53.25 & 80.72 \\
    \%Global Edit Request & 46.75 & 19.28 \\
    \midrule
    \multicolumn{3}{c}{
    \cellcolor[rgb]{0.95, 0.95, 0.95}
    \textbf{Localization of Edit Request (for Edit Request Only)}}\\
    \midrule
    \%User Requests Explicit Localization & 13.25 & 12.94 \\
    \%User Requests Implicit Localization & 86.75 & 87.06 \\
    \midrule
    \multicolumn{3}{c}{
    \cellcolor[rgb]{0.95, 0.95, 0.95}
    \textbf{System Reply: Number of Executable Edits Suggested}} \\
    \midrule
    \%No Suggestion (pure text reply) & 33.97 & 18.38 \\
    \%Single-Suggestion & 39.23 & 72.79 \\
    \%Multi-Suggestion (2+) & 26.79 & 8.82 \\
    \bottomrule
    \end{tabular}
    }
    \caption{Statistics on qualitative differences in conversations between the participant and the system in the global Chat and the localized Comments (based on participants in Study 1).}
    \label{tab:chat_vs_comments}
\end{table}

The Chat and Comment components of InkSync each provide a conversational interface to the participant. Table~\ref{tab:chat_vs_comments} summarizes an analysis of the properties of conversations that occur in the Chat and Comment components.
The analysis is based on all the interaction logs of participants in Study 1 and is based on 60 Chat conversations and 97 Comment conversations. (we did not include the conversations from the baseline \texttt{NoEx Chat} condition.)

First, Chat conversations tend to be much longer -- 8.6 messages exchanged -- compared to Comment conversations -- 3.0 messages exchanged. We manually classify participants' messages into three classes: Edit Request (e.g., ``suggest what to eat in Egypt''), Information Seeking (e.g., ``what is the capital of Argentina''), and Chit-Chat (e.g., ``hi'', ``thanks''). All three classes occur roughly equally in Chat conversations, whereas participant messages in Comment conversations are much more likely to be Edit Requests.

For participant messages that were labeled as Edit Request, we further labeled whether the scope of the requested edit is local in which editing would only be required in a subpart of the document (i.e., ``add a sentence about food''), or global (e.g., ``fix all the typos in the document''). Unsurprisingly, edit requests in Comment conversations are very likely to have a local scope (80\%). The scope is almost evenly split in Chat conversations, indicating that participants use the Chat component to request edits that are both local and global in scope.

For participant messages that requested an edit, we also labeled whether the request explicitly specifies the desired location for the edit (Explicit Localization: ``Add a sentence about food in the first paragraph.'') or not (No Localization: ``Can you add some more ideas for activities for kids in Singapore''). In both conditions, participants predominantly did not specify localization, with only around 13\% of user messages providing an explicit location for the desired edit.

The combination of the last two statistics suggests that in Chat conversations, participants frequently requested local edits without providing explicit localization.

This section so far has described participant messages in the Chat and Comment conversation panes. Regarding the components' responses, we analyze the number of suggested edits averaged across all participants, grouping them by No Suggestion (0), Single-Suggestion (1), and Multi-Suggestion (2+) responses. No suggestions can occur for information-seeking messages, for instance. In the Chat component, the chat is almost equally likely to reply with no suggestions, with one or multiple suggestions. On the other hand, system replies to Comment requests are much more likely to suggest a single edit.

In summary, although both the Chat and Comment components offer a similar conversational interface to the user, we observe important differences in conversations that occur in each component. Chat conversations are longer, with more diverse participant intent, where they request edits of both local and global scope while not usually specifying location explicitly. This causes the system to have to implicitly resolve ambiguity and produce wider variability in the number of suggested edits. By contrast, Comment conversations are shorter, with participants predominantly requesting editing of local scope, and the system frequently suggests a single edit in response, representing a much more focused and efficient localized conversation.

\subsection{Study 1: Factual Accuracy Annotation} \label{app:study1_accuracy}

\begin{table}[]
    \begin{tabular}{llp{5.3cm}p{5.3cm}}
    \toprule
    \textbf{Error Type} & \textbf{Severity} & \textbf{Email Excerpt} & \textbf{Explanation} \\
    \midrule
    \multirow{2}{*}{Overselling} & Minor & This unique, unrepeatable adventure—a paradise waiting just for you—could be missed with just one delay. & Might be overselling. It creates a sense of urgency which can be considered a pushy sales technique. \\
     & Major & New York Schools are creating the future scientists. & A sweeping generalization that all New York Schools are focused on creating future scientists, which may not be true. \\
     \midrule
    \multirow{2}{*}{Persona Error} & Minor & Indulge in croissants, cheese, and escargot. & It's quite uncommon for 6-10 yo kids to enjoy snails. \\
     & Major & Cycling is also a popular option in Singapore. & Given the persona is a retired person with limited mobility, cycling as a mode of transport is inappropriate. \\
     \midrule
    \multirow{2}{*}{Destination Error} & Minor & Also, Paris is budget-friendly so you can see all the sights without breaking your budget. & Paris is known for being one of the most expensive cities in the world, not typically referred to as "budget-friendly". \\
     & Major & Iguazu Falls is a gorgeous tourist attraction to swim and take photos & It's inaccurate because swimming is not allowed at Iguazu Falls due to safety reasons. \\
    \bottomrule     
    \end{tabular}
    \caption{Examples of inaccurate information identified in emails edited by participants in Study 1. We provide an example for each error category (Overselling, Destination-error, Persona-error), and severity (Minor, Major), as well as an explanation for the error.}
    \label{tab:study1_inacurrate_examples}
\end{table}
\begin{table}[]
    \begin{tabular}{ccccccccc}
    & & \multicolumn{6}{c}{\textbf{Study Condition}} & \\
    \cmidrule(r){1-2} \cmidrule(r){3-8} \cmidrule(r){9-9}
    \textbf{Error Type} & \textbf{Severity} & \textbf{Manual} & \textbf{Markers} & \textbf{NoEx Chat} & \textbf{Localized} & \textbf{Chat} & \textbf{4-Comp} & \textbf{Total} \\
    \cmidrule(r){1-2} \cmidrule(r){3-8} \cmidrule(r){9-9}
    Number of Docs & - & 34 & 32 & 29 & 32 & 30 & 35 & 192 \\
    \cmidrule(r){1-2} \cmidrule(r){3-8} \cmidrule(r){9-9}
    \multirow{2}{*}{Overselling} & Minor & 5 & 3 & 1 & 5 & 6 & 4 & 24 \\
     & Major & 0 & 1 & 1 & 0 & 1 & 0 & 3 \\
    \multirow{2}{*}{Persona Error} & Minor & 1 & 3 & 4 & 1 & 6 & 3 & 18 \\
     & Major & 0 & 0 & 2 & 0 & 2 & 2 & 6 \\
    \multirow{2}{*}{Destination Error} & Minor & 3 & 1 & 1 & 2 & 3 & 1 & 11 \\
     & Major & 3 & 0 & 2 & 5 & 2 & 1 & 13 \\
    \cmidrule(r){1-2} \cmidrule(r){3-8} \cmidrule(r){9-9}
    \multirow{2}{*}{Total} & Minor & 9 (26\%) & 7 (22\%) & 6 (21\%) & 8 (25\%) & 15 (50\%) & 8 (23\%) & 53 (28\%) \\
     & Major & 3 (9\%) & 1 (3\%) & \textbf{5 (17\%)} & \textbf{5 (16\%)} & \textbf{5 (17\%)} & 3 (9\%) & 22 (12\%) \\
     \bottomrule
    \end{tabular}
    \caption{Summary of identified factual errors in the 192 emails edited by participants in Study 1. Errors are broken down by study condition and severity of the error. Note that this count is a lower-bound, as some factual errors might not have been detected during our two-stage annotation process.}
    \label{tab:study1_accuracy_breakdown}
\end{table}

Similar to the analysis we performed for the Objective A/B calculations, we used a two-stage approach to identify the factual errors in the document. In the first stage, we prompted GPT-4 to generate a list of factual errors in each of the edited emails, a list we reviewed and updated in the second manual stage.

We note that for this annotation, GPT-4 tended to produce a large number of false positives, with only one in five initially detected errors being confirmed manually as an error, indicative of the difficulty of factual inaccuracy detection for LLMs.

False negatives are also a potential issue: since we only reviewed potential errors from the list generated by GPT-4, our annotation does not contain any error not originally listed by GPT-4. We therefore encourage the reader to see our analysis as a lower bound on the number of errors present in the emails, with additional unidentified errors possibly present. Since we used the same annotation procedure for all conditions, we expect the number of unidentified errors to be similar across conditions, and therefore not affect relative comparisons.

Table~\ref{tab:study1_inacurrate_examples} lists an identified example of each error category and severity, and Table~\ref{tab:study1_accuracy_breakdown} summarizes the results of the annotation, broken down by interface conditions. Across conditions, 28\% of documents contain a minor factual error, and 12\% contain a major error. Major errors vary slightly based on the condition: 14.8\% of documents edited in a condition with a conversational interface (NoEx Chat, Localized, Chat, or 4-Comp) contain a major error, compared to 6\% for non-conversational interfaces (Manual, Markers). We therefore find evidence that recommendations made by the system in conditions with conversational components lead to a higher likelihood of the participant integrating a major factual error in the email.

\subsection{Study 1: Component Response Time Analysis} \label{app:study1_response_time}

\begin{table}[]
    \centering
    \begin{tabular}{lrccc}
         & & \multicolumn{3}{c}{Response Time (sec)} \\
        \cmidrule(l){3-5}
        Component & N & 10th Percentile & Average & 90th Percentile \\
        \midrule
        \symbolimg{figures/icons/marker_symbol.png} Markers & 801 & 8.2 & 15.8 & 23.8 \\
        \symbolimg{figures/icons/chat.png} Chat & 757 & 1.6 & 6.5 & 14.5 \\
        \symbolimg{figures/icons/comment.png} Comment & 247 & 2.0 & 6.1 & 11.5 \\
        \symbolimg{figures/icons/brainstorm.png} Brainstorm & 193 & 0.0 & 7.8 & 15.3 \\
        \bottomrule
    \end{tabular}
    \caption{Statistics on response time of LLM-based components of InkSync, for participants of Study 1. For each component, we report the number of logged interactions (N), and the response time in seconds: (10th percentile, average and 90th percentile).}
    \label{tab:response_time}
\end{table}

In Table~\ref{tab:response_time}, we report on the average response time of the four LLM-based edit-suggesting components of InkSync. The Chat, Comment, and Brainstorm components typically respond to author queries in 6-8 seconds, while the Markers component responds in 16 seconds. The slower nature of the Markers components is less problematic, as the component predominantly runs in the background, and suggests edits to the user proactively.

The development of LLMs is an active area of research, and we expect that LLMs with equal capabilities of GPT-4 but of smaller size will lead to lower response time, which will reduce component latency without lowering the quality of suggested edits.

\subsection{Study 2: Example System-Generated Inaccuracies} \label{app:study2_inaccuracies}

\begin{figure}
    \centering
    \includegraphics[width=0.95\textwidth]{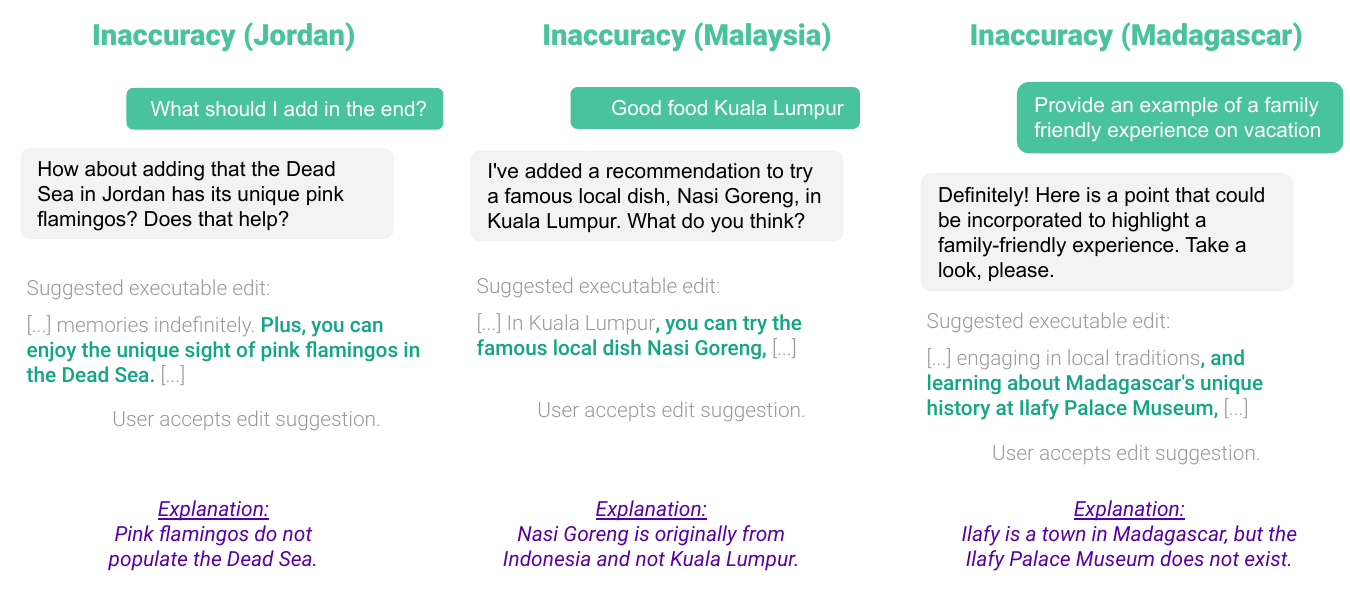}
    \caption{Examples of inaccurate executable edits suggested to participants in Study 2 by the Chat component under the perturbed prompt. Initial participant query in the green box, followed by the plain text reply by the Chat component (grey box), and the component's suggested executable edit. We add an explanation (purple) to signify why the edit is inaccurate.}
    \label{fig:study2_example_inaccuracies}
\end{figure}

In Study 2, we implement a perturbed prompt for the Chat component, that encourages the LLM to suggest factually inaccurate information in the edit suggestions. Figure~\ref{fig:study2_example_inaccuracies} lists an example of an inaccurate suggestion made by the Chat component under the perturbed prompt to participants in Study 2, for each of the three destinations in the study.

As designed, the recommendations include accuracies that could be verified through a Google search (e.g., where is Nasi Goreng from?), while not necessarily obvious to a US-based knowledge worked (e.g., claiming that Sushi was invented in New York).

\subsection{Statistical Significance Analysis of Main Findings} \label{app:stats_significance}

\begin{figure*}
    \centering
    \begin{minipage}[t]{0.45\textwidth}
        \subfloat[Study 1: Objective A/B differences (Section~\ref{sec:study1_evaluation_goal})]
        {\label{fig:stat_sig_objectives} \includegraphics[width=\textwidth]{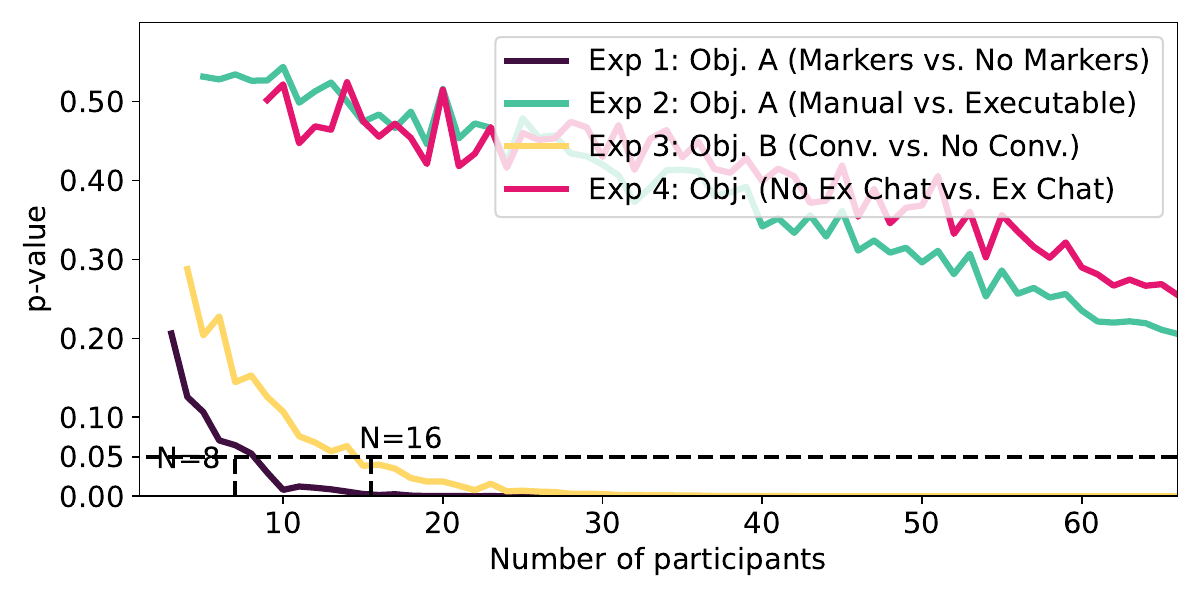}}
    \end{minipage}\hfill
    \begin{minipage}[t]{.45\textwidth}
        \subfloat[Study 1: Editing Efficiency (Section~\ref{sec:study1_evaluation_efficiency})]
        {\label{fig:stat_sig_efficiency} \includegraphics[width=\textwidth]{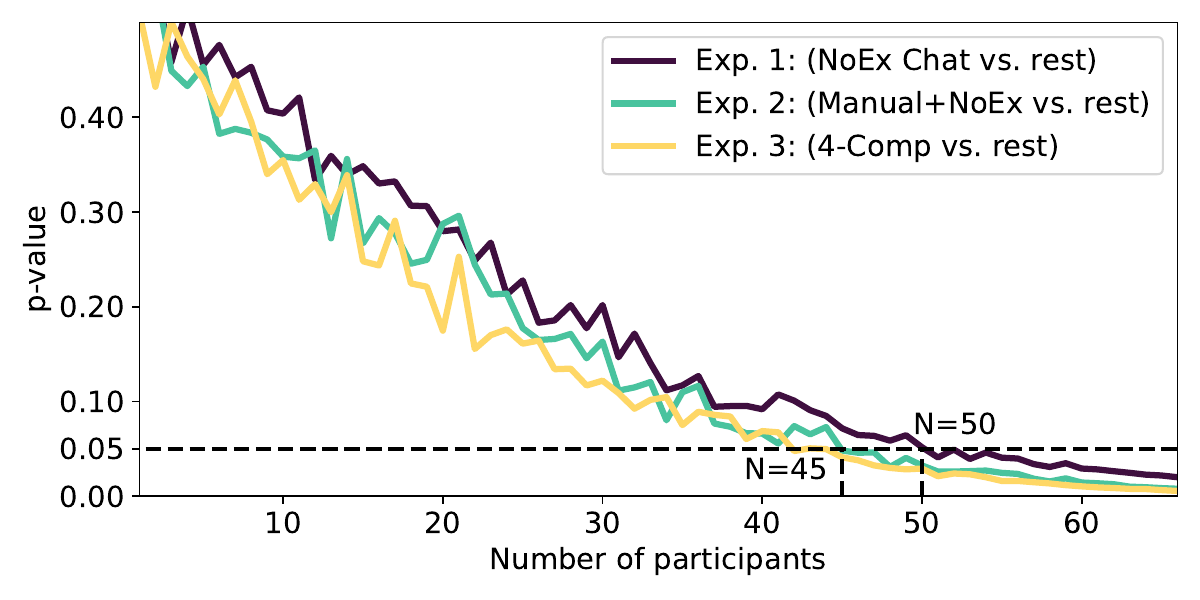}}
    \end{minipage}\hfill
    \vspace{0.7cm}
    \begin{minipage}[t]{.45\textwidth}
        \subfloat[Study 1: Ranking Results (Section~\ref{sec:study1_evaluation_preference})]
        {\label{fig:stat_sig_ranking} \includegraphics[width=\textwidth]{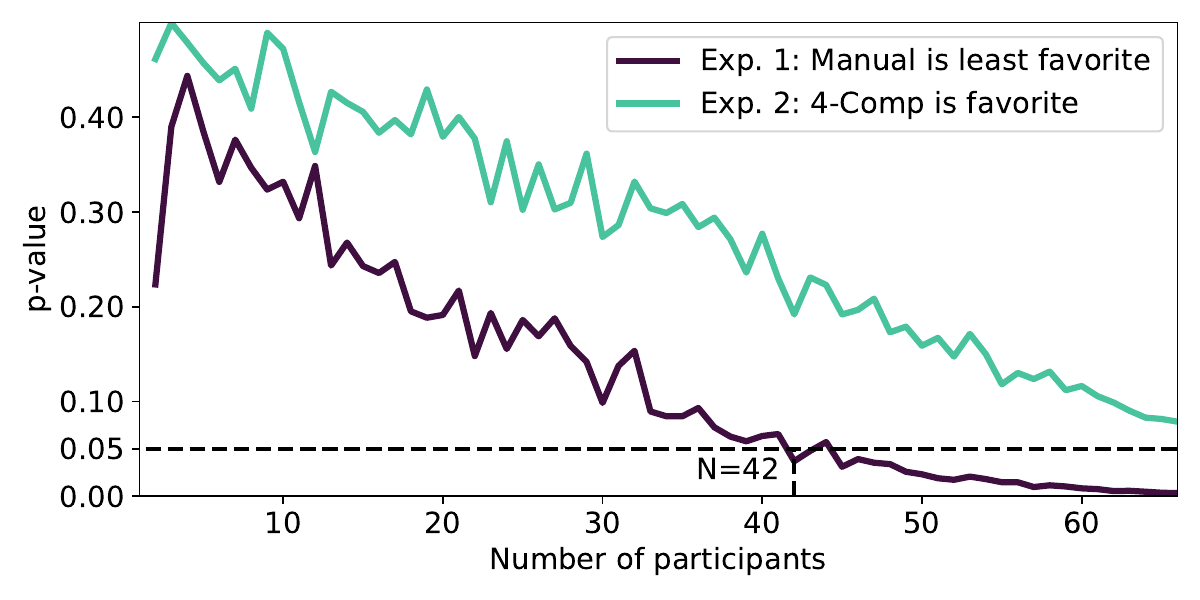}}
    \end{minipage} \hfill
    \begin{minipage}[t]{.45\textwidth}
        \subfloat[Study 2: Factual Error Prevention and Detection (Section~\ref{sec:study2_evaluation_inaccuracy})]
        {\label{fig:stat_sig_study2} \includegraphics[width=\textwidth]{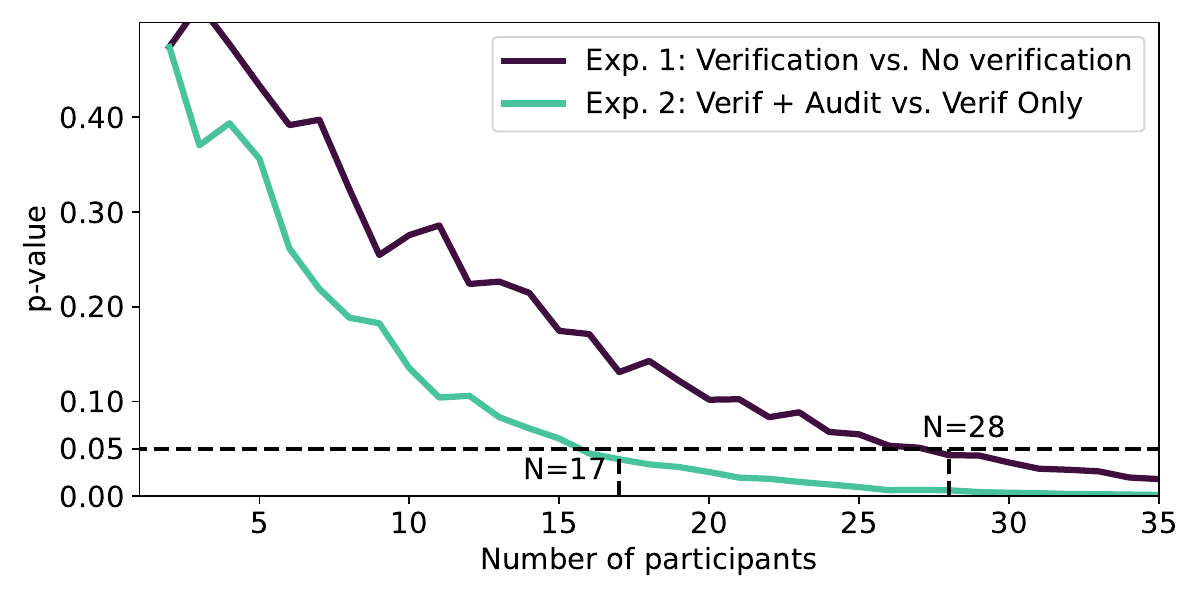}}
    \end{minipage}
    \caption{Results of the bootstrap re-sampling analysis for main findings of the usability studies we conducted. Although each plot contains all statistical significance testing of a subsection, we note that each experiment is independent, and was grouped to save space.}
    \label{fig:stat_sig}    
\end{figure*}

In order to establish the statistical significance of the findings of the usability studies (Section~\ref{sec:study1}-\ref{sec:study2}), we perform statistical tests with the entire participant population, as well as bootstrap re-sampling \cite{efron1982jackknife} (varying the number of participants through sub-sampling) to determine the minimum number of participants required to achieve statistical significance ($p < 0.05$). When performing multiple comparisons relying on the same data, we use the Bonferroni correction \cite{bonferroni1935calcolo} to adjust the significance level accordingly.

Figure~\ref{fig:stat_sig} presents the results of the four sets of statistical tests we performed.  (Note that results from multiple studies are placed within individual charts to save space and facilitate comparison; however, each plot is independent of the other and could in principle appear in its own chart.)

Figure~\ref{fig:stat_sig_objectives} includes four analyses: (1) Experiment 1: whether the inclusion of the Marker component leads to an improvement in Objective A (significant with 8 participants or more), (2) Experiment 2: whether any executable edit component leads to improvements over the manual interface (not statistically significant), (3) Experiment 3: whether the inclusion of a conversational component (Chat or Comment) leads to an improvement in Objective B (significant with 16 participants or more), (4) Experiment 4: Comparing the NoEx Chat and Chat conditions, whether there is a statistical significant improvement on both scores (not significant).

Figure~\ref{fig:stat_sig_efficiency} includes three analyses: (1) Experiment 1: whether the NoEx Chat condition leads to the least amount of editing in the five-minute editing sessions (significant starting with 50 participants), (2) Experiment 2: whether the conditions without executable edits (Manual, NoEx Chat) lead to slower editing than conditions with executable edits (Markers, Chat, Comment, 4-Comp), which is significant with 45 participants or more, and (3) Experiment 3: whether the 4-comp condition leads to faster editing than the single-component conditions (Markers, Chat, Comment), which is significant with 45 participants or more.

Figure~\ref{fig:stat_sig_ranking} includes two analyses: (1) Experiment 1: whether the Manual interface received the lowest ranking in terms of participant preferences (significant with 42 participants or more), and (2) Experiment 2: whether the 4-Comp condition was most favored by participants (not significant, with 65 participants $p=0.079$.

Figure~\ref{fig:stat_sig_study2} includes two analyses based on Study 2: (1) Experiment 1: whether the inclusion of the Warn and Verify components leads to a larger proportion of avoided factual errors (Step 1 vs. Step 2), which is significant with 28 participants or more, (2) Experiment 2:  whether performing an audit leads to additional error reduction, compared to edit-time-only verification (which is significant with 17 participants or more).

\end{document}